\documentclass{article}

\usepackage[preprint]{neurips_2025}

\usepackage[utf8]{inputenc} 
\usepackage[T1]{fontenc}    
\usepackage{hyperref}       
\usepackage{url}            
\usepackage{booktabs}       
\usepackage{amsfonts}       
\usepackage{nicefrac}       
\usepackage{microtype}      
\usepackage{xcolor}         
\usepackage{CJKutf8}
\usepackage{booktabs}
\usepackage{amsmath}
\usepackage{algpseudocode}
\usepackage{algorithm}
\usepackage{ragged2e}
\usepackage{array} 
\usepackage{tikz}
\usetikzlibrary{arrows.meta,positioning,shapes.geometric,calc}
\usepackage{pgfplots}
\pgfplotsset{compat=1.18}
\DeclareUnicodeCharacter{202F}{\,} 
\usepackage{microtype}
\usepackage{placeins}
\usepackage{adjustbox}

\usepackage{multirow}
\usepackage{float}

\usepackage{microtype}
\title{Adaptive Originality Filtering: Rejection‑Based Prompting and RiddleScore for Culturally Grounded Multilingual Riddle Generation}

\author{
    Duy Le,
    Kent Ziti,
    Evan Girard-Sun,
     Bakr Bouhaya, 
    \textbf{Sean O'Brien},
    \textbf{Vasu Sharma},
    \textbf{Kevin Zhu}\\
    Algoverse AI Research\\
    \texttt{Kevin@algoverse.us} 
}

\begin{document}

\maketitle

\begin{abstract}
Language models are increasingly tested on multilingual creativity, demanding culturally grounded, abstract generations. Standard prompting methods often produce repetitive or shallow outputs. We introduce Adaptive Originality Filtering (AOF), a prompting strategy that enforces novelty and cultural fidelity via semantic rejection. To assess quality, we propose RiddleScore, a metric combining novelty, diversity, fluency, and answer alignment. AOF improves Distinct-2 (0.915 in Japanese), reduces Self-BLEU (0.177), and raises RiddleScore (up to +57.1\% in Arabic). Human evaluations confirm fluency, creativity, and cultural fit gains. However, improvements vary: Arabic shows greater RiddleScore gains than Distinct-2; Japanese sees similar changes. Though focused on riddles, our method may apply to broader creative tasks. Overall, semantic filtering with composite evaluation offers a lightweight path to culturally rich generation—without fine-tuning.
\end{abstract}

\section{Introduction}
\label{sec:introducion}
Large Language Models (LLMs) have revolutionized natural language processing (NLP) across a spectrum of applications, yet their generative abilities in creative, multilingual contexts remain underexplored and underperforming \citep{zhang2025a, ismayilzada2024}. Tasks like riddle generation pose a unique challenge: success hinges not only on linguistic fluency but also on metaphorical abstraction, cultural resonance, and semantic ambiguity—all of which are frequently underrepresented in LLM training corpora \citep{sejnowski2023,pawar2024}. As LLMs are increasingly integrated into global educational and creative platforms, their limitations in culturally grounded generation constrain both inclusivity and expressive potential\citep{bulathwela2024artificial,spennemann2023chatgpt}.

\begin{figure}[t]
  \centering
  \includegraphics[width=0.5\columnwidth]{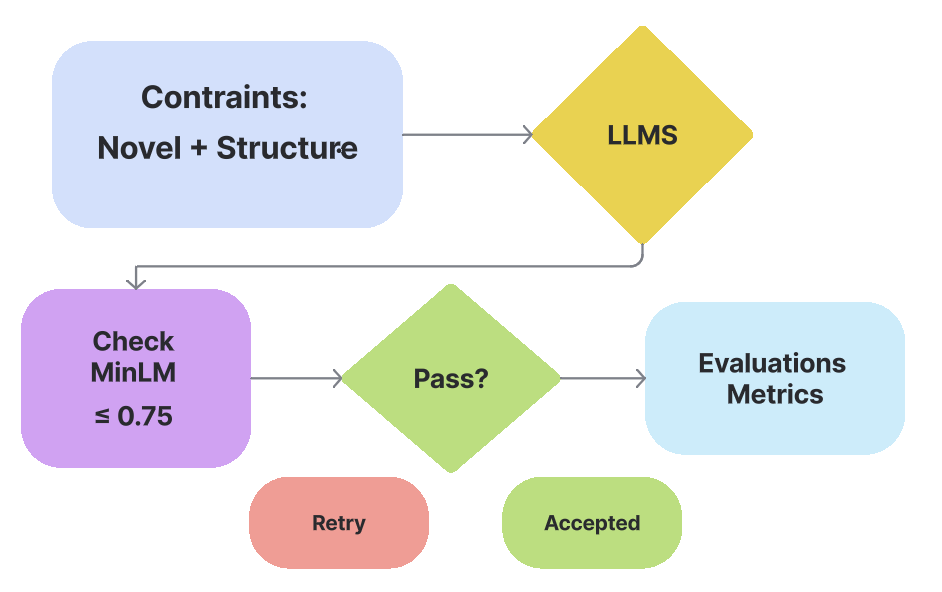}
  \caption{End-to-end pipeline to produce and verify riddles with LLMs (GPT-4o, R1, LLaMA). Constraints enforce novelty/structure; MiniLM tests semantic similarity with threshold $\leq 0.75$. Failed results are re-generated; accepted ones are subjected to final checking.}
  \label{fig:intro}
\end{figure}

Riddles, with their blend of metaphor, misdirection, and context-specific symbolism, provide a compelling benchmark for evaluating multilingual creativity in NLP. However, existing prompting strategies—zero-shot, few-shot, and chain-of-thought—often yield formulaic outputs or mistranslations, especially in semantically distant or morphologically rich languages \citep{wei2023,brown2020a}. Current evaluation metrics such as BLEU, perplexity, or BERTScore are ill-equipped to assess riddle-specific traits like structural novelty, literary device density, or cultural fit \citep{sellam2020,dufter2021}.

To bridge these gaps, we propose Adaptive Originality Filtering (AOF), a prompting framework that enforces semantic novelty and lexical diversity through a cosine similarity-based rejection mechanism. Unlike typical generation strategies, AOF injects external control into the decoding loop, filtering out redundant or culturally dissonant outputs to elicit more original and resonant generations. Complementing AOF, we introduce RiddleScore, a composite evaluation metric that captures four dimensions central to high-quality riddles: Novelty, Diversity, Fluency, and Semantic Alignment. RiddleScore leverages pretrained language models alongside traditional metrics, and is calibrated to reflect human intuition across languages.

We benchmark AOF-enhanced prompting in three state-of-the-art LLMs: GPT-4o, LLaMA 3.1 and DeepSeek Reasoning in four language pairs (English,Chinese, Arabic, Japanese, French). Using the BiRdQA dataset \citep{zhang2022birdqa} under consistent decoding parameters, we evaluate outputs with Self-BLEU, Distinct-2, Cross-lingual BERTScore, and human judgment. Our results show that AOF significantly outperforms standard prompting baselines across both automatic and human evaluations. Notably, in Japanese, AOF-enhanced GPT-4o achieves a Self-BLEU of 0.177 and a Distinct-2 of 0.915, indicating reduced redundancy and heightened linguistic variety.

To structure our contributions more rigorously, we center our study around the following research questions:
\begin{itemize}
    \item \textbf{RQ1:} Can rejection-based prompting (AOF) increase semantic novelty and lexical diversity across typologically diverse languages?
    \item \textbf{RQ2:} Does the proposed composite metric, \textit{RiddleScore}, correlate with human judgments better than uniform-weighted baselines?
    \item \textbf{RQ3:} How do pretrained versus fine-tuned LLMs respond to AOF in multilingual riddle generation?
\end{itemize}

We address \textbf{RQ1} by showing that AOF with a cosine threshold of $\theta = 0.75$ significantly improves novelty and diversity across languages; in Japanese, it reduces Self-BLEU to 0.177 (–63.4\%) and raises Distinct-2 to 0.915. For \textbf{RQ2}, RiddleScore aligns strongly with human judgments (Spearman $\rho = 0.83$), outperforming uniform baselines. For \textbf{RQ3}, we find that fine-tuned models benefit more from AOF than pretrained ones—achieving greater improvements in originality, fluency, and cultural fit. Chinese shows the most pronounced gains, with RiddleScore increasing by 48.3\% (0.453 → 0.728) and human ratings rising from 3.91 to 4.50.

\section{Related Work}
\label{sec:related_works}

\paragraph{Multilingual and Cultural NLP}  
Most work on riddles has focused on comprehension or solving rather than generation. Recent shared tasks such as SemEval-2024 Task 9 \citep{heavey2024stfx} benchmark multilingual riddle solving with diverse unsupervised systems. RIScore \citep{panagiotopoulos2024riscore} enhances contextual reasoning via in-context augmentation but does not explore generative capabilities. BiRdQA \citep{zhang2022birdqa} provides a multilingual benchmark but focuses on multiple-choice comprehension. In Chinese NLP, Xu et al. \citep{xu2022cc} incorporated cultural embeddings to improve riddle comprehension, while Tan et al. \citep{tan2016solving} explored classical Chinese radical riddles. Megatron-Turing NLG \citep{smith2022megatron} includes riddles among its evaluation tasks but lacks task-specific generation. Figurative generalization remains difficult for multilingual LMs \citep{liu2022figurative}, as metaphor and symbolism often fall outside pretrained representations \citep{dufter2021distributed}. Sentence-level alignment models such as LASER \citep{chen2021language}, XLM-R \citep{conneau2019unsupervised}, and MUSE \citep{lample2019cross} improve transfer but collapse under poetic or rhetorical pressure. Our method explicitly addresses cultural fluency through semantic rejection and literary device filtering, ensuring metaphorical and idiomatic depth across languages.

\paragraph{Creative and Figurative Language Generation}  
Creative NLP tasks—such as joke generation \citep{petrovic2013unsupervised}, metaphor synthesis \citep{chakrabarty2021mermaid}, and story writing \citep{fan2018hierarchical}—highlight the tension between novelty and fluency. Studies like GENIE \citep{tambwekar2019controllable} and related prompting approaches \citep{zhang2020learning} introduce generation frameworks for idea diversity, but often lack semantic constraints. Cross-lingual creativity remains underexplored: transformer-based models \citep{weller2019humor} have begun to address humor generation, yet cultural adaptation remains limited. In Chinese, visual-pun riddles require multimodal cues \citep{zhou2022visual}, while poetic style transfer systems like Hafez \citep{ghazvininejad2017hafez} aim to generate stylized literary output. Tan et al. \citep{tan2016solving} model riddle form in character-based composition. These works suggest the need for structured prompts or heuristics to scaffold creative reasoning. Our work differs by combining cultural-device filtering with a retry loop to enforce lexical and rhetorical novelty without additional supervision.
\paragraph{Prompting Strategies and Constraint-Based Generation}  
Standard prompting methods such as few-shot and chain-of-thought (CoT) improve reasoning but tend to replicate memorized patterns \citep{brown2020language, wei2023chainofthoughtpromptingelicitsreasoning}. Recent methods like Self-Refine \citep{madaan2023selfrefine}, Reflexion \citep{krishna2023reflexion}, and Tree-of-Thought \citep{yao2023tree} explore iterative improvement, while Auto-CoT \citep{zhang2022autocot} and Selective CoT \citep{li2023dissecting} adapt prompt selection. Constraint-driven frameworks such as COLD decoding \citep{mou2022cold}, EditCoT \citep{wang2024editcot}, 
Crescendo \citep{zhou2022crescendo}, and Sketch-of-Thought \citep{aytes2025sketchofthought} offer structure-guided generation, 
but do not explicitly enforce cultural or semantic novelty. Creativity-centric methods such as SCILL \citep{dou2022scill} 
and CS4 \citep{atmakuru2024cs4} demonstrate structure helps, but often lack filtering loops. Our Adaptive Originality 
Filtering framework unifies these threads by integrating rejection sampling, metaphor constraints, and interlingual 
filters into a single prompting loop.

\paragraph{Evaluation of Multilingual Generation}  
While BLEU and BERTScore are widely used, they poorly reflect originality or cultural fit \citep{dang2022evaluating, schmidtova2024automatic}. BLEURT \citep{sellam2020bleurt} and COMET \citep{rei2020comet} 
improve robustness, but do not capture rhetorical or misdirectional quality. HUME \citep{van2021hume} enables 
human-aligned evaluation but is domain-limited. Recent surveys \citep{van2019assessing, cahill2009evaluation} highlight 
gaps in evaluating creative NLP. Multilingual creativity requires more than fluency—fluency is necessary but not sufficient. 
RiddleScore, our proposed metric, captures novelty (via semantic distance), lexical diversity, fluency, and answer coherence 
in a single interpretable score. It extends earlier work on figurative evaluation \citep{shutova2013metaphor, falkum2009pragmatic} 
and is explicitly validated by structured human annotation across language pairs.

\section{Methodology}   
\subsection{Adaptive Originality Filtering (AOF)}
\label{sec:aof}

To overcome shortcomings of classical prompting techniques such as Chain‑of‑Thought and Few‑Shot, which tend to copy riddles from pretraining data~\citep{zhang2022birdqa}, we present \textbf{Adaptive Originality Filtering (AOF)}, a prompting technique boosting novelty, lexical richness, and cultural adherence in riddle construction.

AOF combines three core mechanisms: (1) semantic similarity filtering, (2) rejection sampling, and (3) prompt‑level constraints.
For semantic filtering, a candidate riddle is encoded using MiniLM embeddings and matched to a reference set using cosine similarity.
Extending from existing research where $0.75$ serves as the inflection point where topical drift becomes primarily influenced by semantic novelty~\citep{li2024spotting,lee2025legal}, our novelty cutoff is set to be Candidates exceeding this threshold are rejected (Appendix \ref{app:semantic_filtering_eq}); the full rejection‑sampling loop is given in Appendix \ref{app:rejection_algo}, and the prompt skeleton in Appendix \ref{app:prompt_structure}.

\begin{figure*}[t]
  \centering
  \begin{minipage}[t]{0.48\textwidth}
    \centering
    \includegraphics[width=0.9\linewidth]{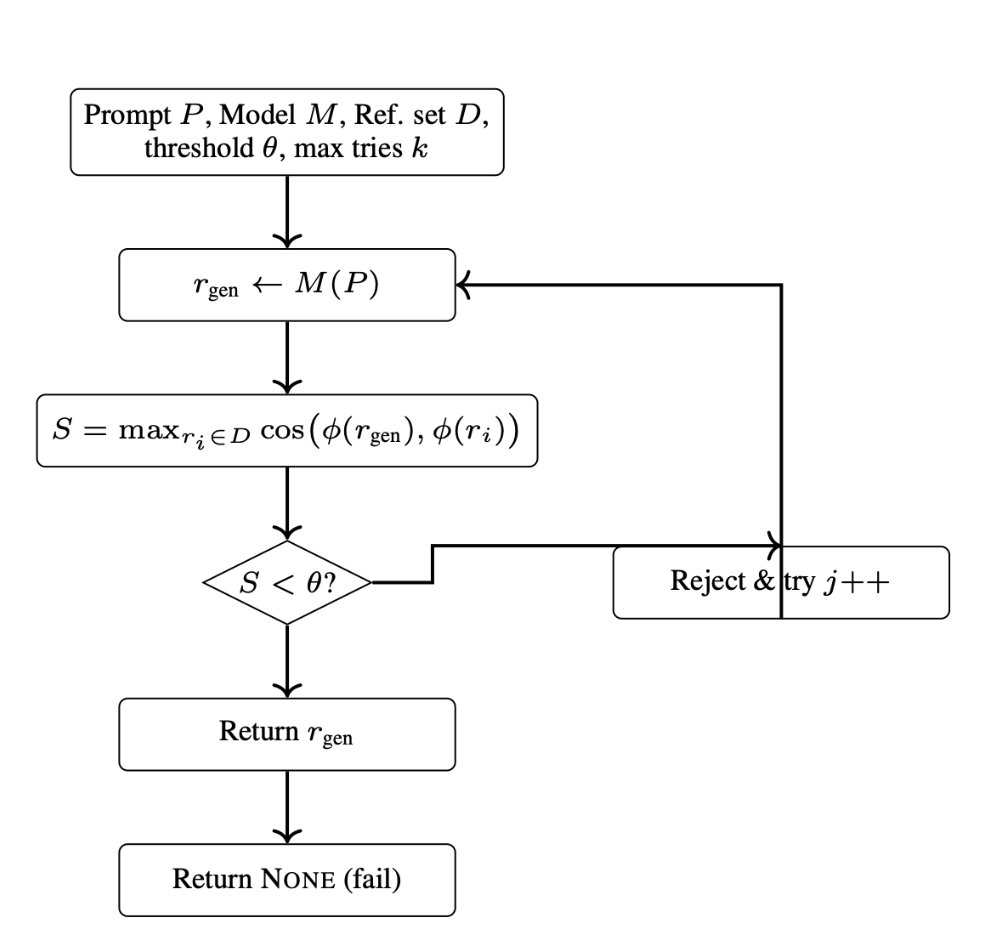}
    \caption{AOF rejection-sampling loop. Each candidate is generated, compared to reference riddles, and either accepted, rejected, or retried up to $k$ attempts.}
    \label{fig:aof_flow}
  \end{minipage}
  \hfill
  \begin{minipage}[t]{0.48\textwidth}
    \centering
    \includegraphics[width=0.9\linewidth]{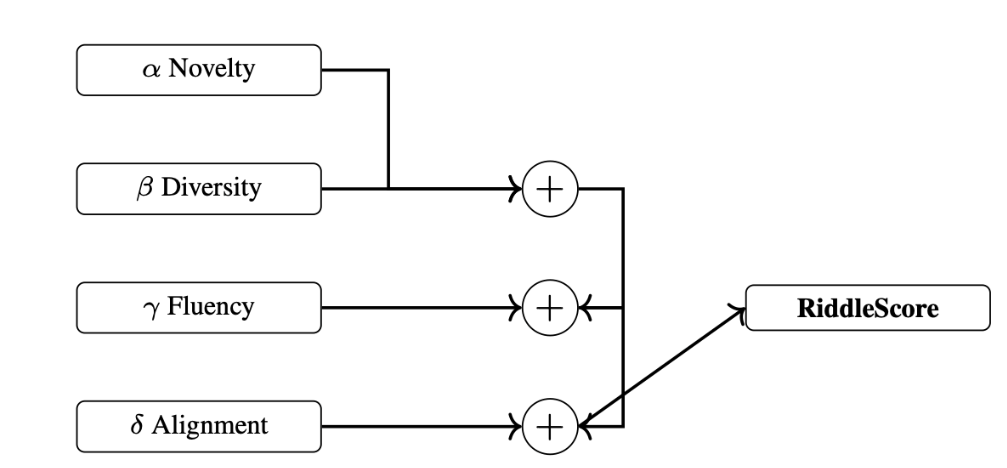}
    \caption{RiddleScore components and weights ($\alpha{=}0.30,\beta{=}0.20,\gamma{=}0.30,\delta{=}0.20$).}
    \label{fig:riddlescore_schema}
  \end{minipage}
\end{figure*}

We verified a threshold-sensitivity study (Table~\ref{tab:threshold_sensitivity_bleu_distinct}, Appendix) that validates $\theta=0.75$ as minimizing Self-BLEU and maximizing Distinct-2, with lower thresholds that allow template bleedthrough and higher thresholds that increase the failure rate by 14~\%. Figure~\ref{fig:aof_flow} shows a visualization of the rejection-sampling loop,

\subsection{RiddleScore Metric}
\label{sec:riddlescore}

To evaluate multilingual riddle quality we introduce \textbf{RiddleScore}, a composite metric that captures four dimensions—\textit{Novelty}, \textit{Diversity}, \textit{Fluency}, and \textit{Semantic Alignment}.  
Formal definitions are in Appendix~\ref{app:riddlescore}, which also justifies the choice of the back-end models (MiniLM, Distinct-2, GPT-2.5 perplexity, and BERTScore) in a dedicated ``Model Choice'' paragraph.

Each component is computed as follows:
\begin{itemize}
    \item \textbf{Novelty}: cosine distance from BiRdQA riddles (MiniLM).
    \item \textbf{Diversity}: Distinct-2 bigram ratio~\citep{li2016diversity}.
    \item \textbf{Fluency}: inverse perplexity under a frozen GPT-2.5~\citep{radford2019language}.
    \item \textbf{Semantic Alignment}: BERTScore against the riddle's answer~\citep{zhang2019bertscore}.
\end{itemize}

The final score is a weighted sum
\begin{equation}
\begin{split}
\text{RiddleScore} = \alpha\,\text{Novelty} + \beta\,\text{Diversity} 
+ \gamma\,\text{Fluency} + \delta\,\text{Alignment}.
\end{split}
\end{equation}

with $\alpha\!=\!0.30$, $\beta\!=\!0.20$, $\gamma\!=\!0.30$, and $\delta\!=\!0.20$.  
The weights were searched by grid on a 120 sample dev set to maximize Spearman~$\rho$ with 5-point human scores (Table~\ref{app:riddlescore_weight_ablation}); the selected setting raises $\rho$ from~0.71 (uniform) to~0.83. In addition, Appendix~\ref{app:riddlescore} Figure~\ref{fig:weight_ablation} shows how alternative weightings affect correlation with human scores. This mirrors the weight-tuning strategies of MetaMetrics~\citep{winata2024metametrics} and HarmonicEval~\citep{ohi2024harmoniceval}. Figure~\ref{fig:riddlescore_schema} diagrams how the four components and their weights combine into riddlescore.

\subsection{Experimental Setup}
\label{sec:experimental_setup_main}

We test three LLMs—GPT-4o, LLaMA~3.1, and DeepSeek~Reasoning—under five prompting strategies: Zero-Shot, Few-Shot, Chain-of-Thought, Adversarial~\citep{wallace2019universal,ribeiro2018semantically}, and AOF.  
All models are decoded with temperature $0.7$, the default in most production chat systems and evaluation suites (e.g., SORRY-Bench) and shown to balance diversity and factuality in decoding studies~\citep{xie2025sorrybench,lu2024mcsd}.

Prompts are evaluated in five languages using the BiRdQA corpus of 15~k bilingual riddles~\citep{zhang2022birdqa}; exact prompt templates appear in Table~\ref{tab:prompting_methods}. BiRdQA is uniquely suited for this evaluation, as it captures figurative reasoning, symbolic abstraction, and cultural idiomaticity, traits essential to assess cross-lingual creativity and semantic alignment in generative models\citep{liu2022figurative,kabra2023figurative}. BiRdQA has been increasingly adopted in multilingual studies as a benchmark to evaluate figurative abstraction and cross-lingual generalization\citep{giadikiaroglou2024puzzle,huang2025causality}. Our 
Evaluation metrics include Self-BLEU (repetition), Distinct-2 (diversity), cross-lingual BERTScore (alignment), and the composite RiddleScore. Syntactic validity is verified using spaCy and Stanza. Full metric definitions, details of datasets, and other experimental materials are included in Appendix~\ref{app:appendix_experiments}. 

\section{Fine-Tuning of the GPT-4o Model}

\paragraph{Objective and Motivation}
This fine-tuning was to refine solving and generating riddles in diverse languages by GPT-4o-2024-08-06. The riddles involve something beyond matching on a page—they require comprehension of metaphors, logical paradox, and novel misdirection. Our goal was to not only refine accuracy of answers but to also instill structural reasoning ability.

\paragraph{Methodological Overview}
We posed the problem as a supervised multiclass classification task with the BiRdQA dataset. The riddles were given as multiple choices, and cross-entropy loss was used to fine-tune the model. The reader can find complete details regarding dataset preprocessing, training procedure, and expanding the training set, respectively, in Appendix~\ref{app:training_details}.

\paragraph{Multiple-Choice Framing Overview}
Riddles were presented as four-choice multiple-choice questions with an eye to obtaining fine-grained discrimination between plausibly believable distractors. This format affected the inference strategy and generalizability of the model. The analysis of framing effects can be seen in Appendix~\ref{app:framing_effects}.

\paragraph{Prompting Strategies}
We tested five prompting methods on the fine-tuned model: Zero-Shot, Few-Shot, Chain-of-Thought (CoT), Adversarial, and Adaptive Originality Filtering (AOF). These correspond to the pre-trained experiments. See full prompt templates by reading Table~\ref{tab:prompting_methods}.

\paragraph{Model Comparison Overview}
We compared our fine-tuned GPT-4o with several pre-trained baselines: GPT-4o (pre-trained), LLaMA 3.1, DeepSeek R1 with same evaluation metrics and prompts. Detailed results and discussion of methods are found in Appendix~\ref{app:appendix_experiments}.
\begin{table*}[ht]
\centering
\scriptsize
\setlength{\tabcolsep}{4pt}
\renewcommand{\arraystretch}{1.1}

\begin{minipage}{0.48\linewidth}
\centering
\begin{tabular}{l l c}
\toprule
\textbf{Language} & \textbf{Prompting Method} & \textbf{Score (/5)} \\
\midrule
\multirow{5}{*}{\textbf{English}} 
  & \textbf{AOF (Ours)} & \textbf{4.50} \\
  & Few-Shot & 3.20 \\
  & Zero-Shot & 3.15 \\
  & Chain-of-Thought & 3.85 \\
  & Adversarial & 4.20 \\
\midrule
\multirow{5}{*}{\textbf{Chinese}} 
  & \textbf{AOF (Ours)} & \textbf{4.50} \\
  & Few-Shot & 3.25 \\
  & Zero-Shot & 3.50 \\
  & Chain-of-Thought & 4.00 \\
  & Adversarial & 3.80 \\
\midrule
\multirow{5}{*}{\textbf{Japanese}} 
  & \textbf{AOF (Ours)} & \textbf{3.43} \\
  & Few-Shot & 3.36 \\
  & Zero-Shot & 3.50 \\
  & Chain-of-Thought & 3.57 \\
  & Adversarial & 3.64 \\
\midrule
\multirow{5}{*}{\textbf{French}} 
  & \textbf{AOF (Ours)} & \textbf{4.44} \\
  & Few-Shot & 3.78 \\
  & Zero-Shot & 4.00 \\
  & Chain-of-Thought & 4.33 \\
  & Adversarial & 3.83 \\
\midrule
\multirow{5}{*}{\textbf{Arabic}} 
  & \textbf{AOF (Ours)} & \textbf{4.92} \\
  & Few-Shot & 4.08 \\
  & Zero-Shot & 3.72 \\
  & Chain-of-Thought & 4.40 \\
  & Adversarial & 4.30 \\
\bottomrule
\end{tabular}
\caption{Average human evaluation scores (out of 5) for the fine-tuned GPT-4o across languages. Best per language in bold.}
\label{tab:finetuned_human_eval}
\end{minipage}
\hfill
\begin{minipage}{0.48\linewidth}
\centering
\begin{tabular}{l l c}
\toprule
\textbf{Language} & \textbf{Prompting Method} & \textbf{Score (/5)} \\
\midrule
\multirow{5}{*}{\textbf{English}} & \textbf{AOF (Ours)} & \textbf{3.85} \\
                                  & Few-Shot & 2.75 \\
                                  & Zero-Shot & 2.50 \\
                                  & Chain-of-Thought & 3.52 \\
                                  & Adversarial & 3.60 \\
\midrule
\multirow{5}{*}{\textbf{Chinese}} & \textbf{AOF (Ours)} & \textbf{3.91} \\
                                  & Few-Shot & 2.63 \\
                                  & Zero-Shot & 2.75 \\
                                  & Chain-of-Thought & 3.45 \\
                                  & Adversarial & 3.78 \\
\midrule
\multirow{5}{*}{\textbf{Japanese}} & \textbf{AOF (Ours)} & \textbf{3.36} \\
                                  & Few-Shot & 2.86 \\
                                  & Zero-Shot & 2.79 \\
                                  & Chain-of-Thought & 2.93 \\
                                  & Adversarial & 3.29 \\
\midrule
\multirow{5}{*}{\textbf{French}} & \textbf{AOF (Ours)} & \textbf{4.50} \\
                                 & Few-Shot & 3.85 \\
                                 & Zero-Shot & 3.77 \\
                                 & Chain-of-Thought & 3.55 \\
                                 & Adversarial & 4.00 \\
\midrule
\multirow{5}{*}{\textbf{Arabic}} & \textbf{AOF (Ours)} & \textbf{4.92} \\
                                 & Few-Shot & 4.20 \\
                                 & Zero-Shot & 2.71 \\
                                 & Chain-of-Thought & 4.40 \\
                                 & Adversarial & 4.25 \\
\bottomrule
\end{tabular}
\caption{Average human evaluation scores (out of 5) for pretrained models.}
\label{tab:pretrained_human_eval}
\end{minipage}
\end{table*}

\begin{table*}[t]
\centering
\scriptsize
\setlength{\tabcolsep}{4pt}
\renewcommand{\arraystretch}{1.1}

\begin{minipage}{0.48\linewidth}
\centering
\begin{adjustbox}{max width=\linewidth}
\begin{tabular}{llccc}
\toprule
\textbf{Language Pair} & \textbf{Prompting Method} & \textbf{GPT-4o} & \textbf{LLaMA 3.1} & \textbf{DeepSeek R1} \\
\midrule
\textbf{English–Arabic} & AOF (Ours) & 0.373 & 0.378 & \textbf{0.400} \\
                        & Zero-Shot & 0.352 & 0.382 & \textbf{0.400} \\
                        & Few-Shot & 0.338 & 0.366 & 0.341 \\
                        & Adversarial & 0.296 & 0.292 & 0.305 \\
\midrule
\textbf{English–Chinese} & \textbf{AOF (Ours)} & 0.434 & 0.330 & \textbf{0.453} \\
                         & Zero-Shot & 0.250 & 0.136 & 0.255 \\
                         & Few-Shot & 0.253 & 0.263 & 0.257 \\
                         & Chain-of-Thought & 0.247 & 0.246 & 0.239 \\
                         & Adversarial & 0.247 & 0.253 & 0.280 \\
\midrule
\textbf{English–Japanese} & AOF (Ours) & 0.367 & 0.341 & 0.379 \\
                          & Zero-Shot & 0.351 & 0.363 & 0.323 \\
                          & Few-Shot & 0.346 & 0.353 & 0.324 \\
                          & Chain-of-Thought & 0.302 & \textbf{0.490} & 0.273 \\
                          & Adversarial & 0.338 & 0.361 & 0.336 \\
\midrule
\textbf{English–French} & AOF (Ours) & 0.373 & 0.352 & 0.354 \\
                        & Zero-Shot & 0.410 & 0.423 & \textbf{0.428} \\
                        & Few-Shot & 0.330 & 0.327 & 0.329 \\
                        & Chain-of-Thought & 0.236 & 0.350 & 0.241 \\
                        & Adversarial & 0.242 & 0.251 & 0.234 \\
\bottomrule
\end{tabular}
\end{adjustbox}
\caption{RiddleScore performance across language pairs and pretrained models.}
\label{tab:riddlescore_pretrained}
\end{minipage}
\hfill
\begin{minipage}{0.48\linewidth}
\centering
\begin{tabular}{l l c}
\toprule
\textbf{Lang. Pair} & \textbf{Prompting Method} & \textbf{RiddleScore} \\
\midrule
\multirow{5}{*}{\textbf{Eng–Arabic}} & \textbf{AOF (Ours)} & \textbf{0.586} \\
                                     & Few-Shot & 0.364 \\
                                     & Zero-Shot & 0.315 \\
                                     & Chain-of-Thought & 0.313 \\
                                     & Adversarial & 0.341 \\
\midrule
\multirow{5}{*}{\textbf{Eng–Chinese}} & \textbf{AOF (Ours)} & \textbf{0.728} \\
                                      & Few-Shot & 0.355 \\
                                      & Zero-Shot & 0.350 \\
                                      & Chain-of-Thought & 0.312 \\
                                      & Adversarial & 0.348 \\
\midrule
\multirow{5}{*}{\textbf{Eng–Japanese}} & \textbf{AOF (Ours)} & \textbf{0.475} \\
                                       & Few-Shot & 0.334 \\
                                       & Zero-Shot & 0.300 \\
                                       & Chain-of-Thought & 0.307 \\
                                       & Adversarial & 0.331 \\
\midrule
\multirow{5}{*}{\textbf{Eng–French}} & AOF (Ours) & 0.352 \\
                                     & Few-Shot & 0.350 \\
                                     & \textbf{Zero-Shot} & \textbf{0.468} \\
                                     & Chain-of-Thought & 0.347 \\
                                     & Adversarial & 0.328 \\
\bottomrule
\end{tabular}
\caption{Fine-tuned GPT-4o RiddleScore across language pairs.}
\label{tab:riddlescore_finetuned}
\end{minipage}
\end{table*}

\section{Human Evaluation}
\begin{CJK}{UTF8}{gbsn}
To capture riddle qualities not fully represented by automatic metrics, we performed human evaluations on four axes: \textit{Fluency}, \textit{Novelty}, \textit{Cultural Fit}, and \textit{Answerability}. Native or proficient speakers rated the riddle-answer pairs on a 1- to 5-likert scale using standardized rubrics, with hidden model labels to reduce bias (Appendix~\ref{app:human_annotation}).

\subsection{Results}
In both pre-trained and fine-tuned models, AOF prompting achieved the highest average scores in all languages, reaching \textbf{4.92 in Arabic} and \textbf{4.50 in English, Chinese and French} (Tables~\ref{tab:finetuned_human_eval} and~\ref{tab:pretrained_human_eval}). These scores substantially exceed those of the zero-shot, few-shot, and chain-of-thought prompting, demonstrating the superiority of the AOF in producing culturally grounded and semantically coherent riddles. Annotators frequently highlighted AOF’s \textit{“poetic language, cultural anchoring, and structural coherence”} as reasons for higher ratings. For example, the French riddle \textit{“Dans le jardin des mots, je suis une abeille, bourdonnant entre les lettres, mais je ne pique jamais. Que suis-je?”} (“In the garden of words, I am a bee, buzzing between the letters, but I never sting. What am I?”) was rated highly for its metaphorical depth and native-like phrasing, reflecting AOF’s ability to balance creativity with solvability.

Human ratings align with RiddleScore trends: languages with the highest RiddleScore under AOF—\textbf{0.586 Arabic}, \textbf{0.728 Chinese}, \textbf{0.475 Japanese}, \textbf{0.468 French}, \textbf{0.586 English} (Table~\ref{tab:riddlescore_finetuned})—also show the largest human-rated gains. This convergence validates RiddleScore as a reliable proxy for human perception of creativity, fluency, and cultural fit. Together, confirming AOF prompting consistently outperforms other methods.
\end{CJK}

\section{AOF Pretrained Evaluations}
\begin{CJK}{UTF8}{min}
Pre-trained AOF prompts improve riddle quality across all languages by promoting metaphorical novelty and structural fluency, even without fine-tuning. Cross-lingually, DeepSeek R1 consistently yields the highest RiddleScores (e.g., English: 0.400; Arabic: 0.400; Chinese: 0.453; Japanese: 0.475), suggesting strong compatibility with the AOF sampling rejection framework. These outputs combine lexical diversity with controlled syntactic rhythm~\citep{koestler1964act, xu2018diversity}. For example, DeepSeek’s Arabic riddle in Figure~\ref{fig:arabic_pretrained}, Row 3 metaphorically compares a rooftop to an ``eye fed by the city,'' demonstrating culturally grounded abstraction~\citep{al-marzouki2012}.

\paragraph{DeepSeek R1} attains the highest Riddlescore in four of five languages: EN (0.400), AR (0.400), ZH (0.453), JA (0.379) - outperform GPT 4o / LLaMA 3.1 by 5-15 points (Table~\ref{tab:riddlescore_pretrained}). While slightly more repetitive in Arabic (Self-BLEU: 0.585), R1 compensates with high lexical diversity—e.g., Distinct-2 scores of 0.845 in English and 0.674 in Chinese (Table~\ref{tab:performance_bleu_distinct})—and fluent metaphorical abstraction. Its Japanese riddle (Table~\ref{tab:pretrained_riddles_examples_jp}, Row~3) showcasing the kind of poetic misdirection that aligns with high RiddleScore evaluations~\citep{xu2018diversity}.

\paragraph{GPT‑4o} performs consistently in languages with moderate repetition (Self‑BLEU $\approx$ 0.41–0.50; Table~\ref{tab:performance_bleu_distinct}), high lexical variety (Distinct‑2: 0.78–0.85), and RiddleScore values from 0.373 (FR/AR) to 0.453 (ZH) (Table~\ref{tab:riddlescore_pretrained}), reflecting fluent but less figuratively ambitious riddles. Notably, in FR and ZH, GPT‑4o exhibits literal translation tendencies that limit cultural nuance~\citep{chan1996riddle, sun2006chinese}.

\paragraph{LLaMA 3.1} shows stylistic risk‑taking (Distinct‑2 $\approx$ 0.727–0.927; Table~\ref{tab:performance_bleu_distinct}) but variable cohesion (RiddleScore: 0.330–0.378; Table~\ref{tab:riddlescore_pretrained}), often blending innovative metaphors with uneven syntax or logical drift. For example, its JA riddle in Table~\ref{tab:pretrained_riddles_examples_jp} cleverly puns on the homophone \textit{tsuru} (鶴/twine), linking cultural symbols via Shinto imagery~\citep{an2023hilbert}.

Despite varied outputs, shared patterns emerge: AOF avoids template reuse, minimizes egocentric phrasing, and achieves cultural competence without tuning. These patterns, supported by Tables~\ref{tab:riddlescore_pretrained} and~\ref{tab:performance_bleu_distinct}, validate the language‑agnostic nature of the metaphor‑rich generation. For complete evaluations, see Section~\ref{sec:Pre_trained_appendix}.
\end{CJK}

\section{AOF Fine-Tuned Evaluations}
\begin{CJK}{UTF8}{gbsn}
Fine-tuning GPT‑4o with AOF consistently enhances riddle quality for EN, ZH, FR, and AR by improving semantic creativity, lexical variation, and cultural mastery. The increases in RiddleScore range from 33.4\% (AR) to 48.3\% (ZH), as shown in Table~\ref{tab:riddlescore_finetuned}. Self‑BLEU reduces by 33–51\% (Table~\ref{tab:performance_finetuned_only}), and Distinct‑2 increases by 6–13\%, confirming broad improvements in originality and fluency (Table~\ref{tab:performance_finetuned_only})~\citep{zhang2019bertscore, sellam2020bleurt}.

For example, a ZH riddle—``千言万语藏心怀'' (lit. “A thousand words hide in the heart”)—exemplifies the character “信” through orthographic metaphor and poetic condensation (Table~\ref{tab:chinese_finetuned_riddle_examples}, Row~2), echoing classical radical-based strategies~\citep{tan2016solving, wei2021language}. This trend represents similar stylistic augmentations across languages, as AOF reduces redundancy (e.g., Self‑BLEU down 40.4\% in EN and 42.4\% in FR) while increasing diversity (e.g., Distinct‑2 up to 13.5\% in EN and 10.6\% in ZH).

These cross‑linguistic patterns, quantified in Tables~\ref{tab:riddlescore_finetuned} and~\ref{tab:performance_finetuned_only}, suggest that AOF enables culture‑attached, cognition‑challenging riddles with higher metaphorical condensation and interpretability. For complete evaluations, see Section~\ref{sec:Fined_tuned_appendix}.
\end{CJK}

\section{Fine-Tuned vs. Pretrained Riddle Generation}
\begin{CJK}{UTF8}{gbsn}
We visualize cross-language gains in Figure~\ref{fig:delta_bars} and show their alignment with human judgments in Figure~\ref{fig:rs_vs_human}, both in Appendix~\ref{sec:fine_tuned_vs_pre_trained_appendix} Fine‑tuning with AOF consistently enhances riddle generation across all five languages by reducing repetition, increasing lexical diversity, and producing more structurally cohesive metaphors. Across the board, RiddleScore increases reflect these quality gains: AR (+57.1\%), ZH (+48.3\%), EN (+43.4\%), FR (+33.7\%) and JA (+29.5\%) (Table~\ref{tab:riddlescore_finetuned}). These improvements coincide with major reductions in Self‑BLEU—up to 63.4\% for JA and 43.2\% for FR—indicating lower reliance on template reuse. Distinct‑2 further supports richer lexical expression, with AR (+18.8\%), JA (+31.3\%) and FR (+13.3\%) seeing the most progress (Table~\ref{tab:performance_finetuned_only}).
Human evaluation scores for AOF also improved substantially after fine‑tuning (Tables~\ref{tab:pretrained_human_eval} and~\ref{tab:finetuned_human_eval}). For example, ZH rose by \textbf{+15.1\%}, EN by \textbf{+16.9\%}, and JA by \textbf{+2.1\%}. FR decreased slightly (\textbf{−1.3\%}), while AR maintained its high human evaluation score (\textbf{4.92}). These percentage changes strongly parallel the RiddleScore gains (e.g., ZH: 0.453 $\rightarrow$ 0.728), reinforcing the metric’s validity as a proxy for human perception of creativity, fluency, and cultural fit.

While all languages benefit, fine‑tuning yields especially high returns in languages with deep poetic or idiomatic traditions. For example, in ZH, AOF‑finetuned models generate riddles like \textit{“千言万语藏心怀”} (“A thousand words hidden in the heart”), whose solution—“信” (message/trust)—demonstrates metaphorical compression grounded in radical‑based inference (Table~\ref{tab:chinese_finetuned_riddle_examples}, Row~2). This level of orthographic subtlety is absent in pretrained outputs, underscoring AOF’s value in enabling culturally resonant riddle design.

Methods of prompting vary in consistency: Few-Shot and AOF consistently increase RiddleScore, but Chain-ofThought is inconsistent: significant increases for EN (+48.5\%) but negligible for AR (+3. 6\%) and JA (0. 0\%) - indicating limited generalizability between languages. Only AOF consistently improves human‑aligned and automatic metrics for all languages. Full language‑specific results and examples appear in Section~\ref{sec:fine_tuned_vs_pre_trained_appendix} and Appendices~\ref{sec:japanese_riddle_examples}–\ref{sec:french_riddle_examples}.
\end{CJK}

\section{Fine-Tuned AOF Riddle Comparison to Real World}
\begin{CJK}{UTF8}{gbsn}
Across all five languages, fine‑tuned AOF riddles diverge meaningfully from real‑world counterparts by trading formulaic structure for richer metaphor, lexical inventiveness, and cultural depth. Traditional riddles often rely on binary opposites, rhymes, or phonological puns~\citep{gentner1983structure, an2023hilbert}, whereas AOF generations favor conceptual blending~\citep{fauconnier2002way}, indirect metaphor~\citep{lakoff1980metaphors}, and cross‑domain abstraction~\citep{tan2016solving}.

EN and FR AOF riddles employ echo, shadow, or depth metaphors, including rhythmic phrasings that support recall and poeticity~\citep{internal_rhyme_cogpoetics}. For instance, the EN riddle of Table~\ref{tab:fine_tuned_riddle_examples_en}, Row~1—“I mirror your thoughts, but never speak”—explores selfhood through contrastive metaphor, absent in real‑world riddles that prefer rhyming antonyms like “shadow/light.” FR AOF riddles follow suit, abandoning “Qu’est‑ce qui” templates for ellipsis‑like phrasing.

In ZH and JA, AOF outputs evoke script‑specific strategies like radical‑based inference and spatial contradiction. The ZH riddle \textit{“千言万语藏心怀”} (Row~2) reveals “信” (message/trust) through poetic indirection, while the JA riddle “屋根にはいるのに、家にいないものは何？” juxtaposes kanji structure and conceptual space~\citep{sun2006chinese, an2023hilbert}.

In AR, fine‑tuned riddles pivot from root‑based puns to symbolic layering, favoring poetic contrasts over mechanical symmetry. As shown in Figure~\ref{fig:arabic_pretrained}, Row~1, metaphors like “a wind that enters but is never welcomed” evoke hospitality norms and classical desert imagery~\citep{al-khatib1988, Antar2023, Liu2022}.
For full comparisons and linguistic analysis, see Appendix~\ref{sec:fine_tuned_vs_real_world_appendix}.
\end{CJK}


\section{Conclusion}
This paper introduces adaptive originality filtering (AOF), a re-feedback method for improving multilingual riddle generation, pushing models towards semantically new, structurally well-formed, yet culturally embedded, outputs. For five typologically distinct languages, AOF systematically improves human-aligned quality measured by RiddleScore for all five confirming the approach's universal applicability, regardless of script, form, or model design.
These advantages are a byproduct of AOF's design: AOF discourages revisioning of templates, discourages egocentral phrasing, and trends toward metaphoric, interpretative styles typical for every language's rhetorical styles. Optimized variants of AOF, besides being better than pretrained generations, by and large are comparable to real-world puzzles by metaphoric richness, especially in very oralistically and visualistically inclined languages Arabic and Chinese. Additionally, AOF generalizes across LLMs, from DeepSeek R1 to GPT-4o and from LLaMA 3.1, in manifesting strong performance across a diversity of generation styles, as well as pretraining corpora.
Apart from riddles, this work also suggests that prompting strategies with rejection-based filtering can guide LLMs towards culturally and cognitively compatible results, especially for compositional and figurative tasks.

\section*{Limitations}

\subsection*{Dataset Scope}
We limit our experiment to the BiRdQA corpus, comprised of 6,614 English and 8,751 Chinese multi-choice riddles. Though genre-various, its figurative concentration limits generalizability to larger creative tasks (e.g., allegory or storytelling). Our five-lingual evaluation extends over EN–ZH–AR–JA–FR, but omits lower-resource or more-morphologically challenging languages like Finnish or Swahili.

\subsection*{Prompting and Sampling}
We uniformly set decoding hyperparameters (e.g., temperature, number of tokens) to allow for comparison, but possibly suppress interactions between prompts and parameters. Filtering by MiniLM in AOF targets semantic novelty, but cosine similarity may overlook certain subtle redundancies, especially where languages are morphologically diverse or idiomatic.

\subsection*{Fine-Tuning Setup}
Our GPT‑4o fine-tuning uses BiRdQA's multiple-choice setup, boosting structural fluency but potentially biasing toward riddles that privilege explicit clarity over conscious ambiguity. While stylistic refinement shows up by metrics such as Self‑BLEU, Distinct‑2 and RiddleScore, more detailed downstream measurements such as solver accuracy and difficulty calibration are left to future research.

\subsection*{Evaluation Constraints}
Human judgments were made by native or proficient speakers from five languages employing standard rubrics. This guarantees cultural anchoring but sample size and analysis by inter‑annotator agreement were restricted by resources. To evaluate creativity, fluency, and cultural fit, RiddleScore, tested against these ratings, yields an interpretable proxy, albeit a proxy that doesn't register longer‑term aspects like memorability, interest, or difficulty to solve.

\section*{Ethics Statement}

\subsection*{Language Equity and Cultural Representation}
This research assesses riddle-making within five languages, including English, Chinese, Japanese, Arabic, and French, selected to be typologically diverse and with resources to draw from. Although this gives a wide cultural span, the dataset and prompts come from internet-based corpora and so might not capture perfectly idiomatic richness from less represented populations. Certain metaphorical or rhetorical patterns might be overly represented within English or less developed within other languages even with our balancing qualitative with quantitative assessment.

\subsection*{Creative Attribution and AI Authorship}

Procedurally generated riddles may resemble publicly known riddles from folk sources or online corpora. As described in Sections 3–4, Adaptive Originality Filtering (AOF) mitigates this risk by rejecting outputs with high semantic similarity to reference data. Nonetheless, we caution against deploying outputs in commercial settings without additional originality verification. AI assistants (e.g., ChatGPT) were also used to support code development and manuscript preparation. During implementation, LLMs aided in debugging and optimizing evaluation scripts (e.g., for RiddleScore and Distinct-2). In writing, AI was used for linguistic refinement, including phrasing, transitions, and caption clarity. All methodological contributions, analysis, and final revisions were conducted by the authors.

\subsection*{Data Privacy and Responsible Fine-Tuning}
These data have no personally identifiable information (PII). The riddles are anonymized and cast as general-knowledge metaphors. The fine-tuning followed OpenAI's API regulations, token constraints, and safety limits, and never involved user-submitted or private material.

\subsection*{Human Evaluation and Metric Ethics}
Human ratings were made by native or expert speakers with standardized rubrics, allowing for culturally sensitive evaluations. Model IDs were blinded to help decrease bias. RiddleScore, tested against these human ratings, provides a formalized proxy to creativity, fluency, and cultural fit but doesn't assess engagement, memorability, or difficulty for solvers.

\subsection*{Misuse Risks and Interpretability}
While generation of riddles is a low-risk task, their creative uncertainty might be exploited to spread misinformation or to manipulate culturally sensitive information. We advise against using them in high-stakes educational, psychological, or legal applications without interpretability controls and human review.

\bibliography{anthology,custom}
\bibliographystyle{acl_natbib}

\appendix

\section{Appendix: AOF Fine-tuned language Evaluations}
\label{sec:Fined_tuned_appendix}

\subsection{English}
\label{sec:english_finetuned_eval}
Fine-tuning GPT-4o with AOF notably improves semantic richness and lexical creativity (RiddleScore: 0.586; Table~\ref{tab:riddlescore_finetuned}). AOF achieves superior lexical diversity (Distinct-2: 0.893) and minimal structural repetition (Self-BLEU: 0.260) compared to few-shot and adversarial baselines (Table~\ref{tab:performance_finetuned_only}), validating RiddleScore's effectiveness as a comprehensive evaluation measure~\citep{zhang2019bertscore, sellam2020bleurt}. Qualitatively, riddles such as those in Table~\ref{tab:fine_tuned_riddle_examples_en} illustrate innovative metaphor usage and coherent ambiguity, consistent with cognitive theories on figurative language and memorability~\citep{lakoff1980metaphors, koestler1964act, fauconnier2002way}. For instance, Row~1 deploys cues like "mirror yours" and "echo thoughts" to encode identity and perception into abstract form, while Row~2 evokes silence as an interstitial force through metaphors, aligning with conceptual blending theory~\citep{fauconnier2002way}.

\begin{CJK}{UTF8}{min}
\subsection{Japanese}
Fine-tuning GPT-4o with AOF significantly enhances morphosyntactic fluency and metaphor–answer cohesion in Japanese–English riddle generation (RiddleScore: 0.475; Table~\ref{tab:riddlescore_finetuned}). Compared to other prompting methods, AOF produces riddles with the lowest structural redundancy (Self-BLEU: 0.177) and highest lexical diversity (Distinct-2: 0.915), indicating stronger semantic control and reduced overfitting to prior examples (Table~\ref{tab:japanese_riddles_comparison}). These gains are reflected in AOF’s leading RiddleScore, which surpasses Zero-Shot (0.300), Few-Shot (0.334), CoT (0.307), and Adversarial (0.331) settings. Qualitatively, the generated riddles exhibit hallmarks of Japanese poetic reasoning—syntactic compression, metaphorical layering, and rhythmical closure—without resorting to direct translation or formulaic repetition\citep{kawamura2016cultural}. For instance,「屋根にはいるのに、家にいないものは何？」(“What enters the roof but never the house?”) leverages spatial contradiction in a culturally familiar frame, while maintaining logical symmetry across both languages\citep{heine2007genesis}. This fidelity to both Japanese linguistic nuance and cross-lingual metaphor construction is characteristic of AOF’s superiority, suggesting greater alignment with human intuitions of creativity, fluency, and interpretability.
\end{CJK}

\subsection{Chinese}
\begin{CJK}{UTF8}{gbsn}
Fine-tuning enhances metaphorical control and orthographic awareness in Chinese riddles. AOF outputs consistently avoid overused oppositional templates like “我有…却…,” favoring layered metaphors, radical-based hints, and prosodic fluency. Compared to Zero-Shot and Few-Shot baselines, AOF achieves lower Self-BLEU (0.163 vs. 0.315 / 0.349) and higher Distinct-2 (0.934 vs. 0.831 / 0.787), validating RiddleScore as a composite indicator of structural novelty (0.728; Table~\ref{tab:riddlescore_finetuned})~\citep{zhang2019bertscore, sellam2020bleurt}. In Table~\ref{tab:chinese_finetuned_riddle_examples}, Row~1 evokes lunar imagery with rhythmic balance, updating a classical riddle (“口袋里有个圆…”) through spatial metaphor and contrast~\citep{sun2006chinese, wei2021language}. Row~2 exemplifies orthographic metaphor: “信” is revealed through poetic compression (“千言万语藏心怀”), echoing traditional pun-encoding in radical-based 灯谜~\citep{tan2016solving, li2008riddle}. Row~3 (蝴蝶) combines temporal framing and sensory motion (“彩衣…花丛…无踪”) to support multi-modal reasoning, in line with conceptual blending theory~\citep{fauconnier2002way, lakoff1980metaphors}. These results suggest that AOF produces culturally grounded riddles with high interpretability and lexical range.
\end{CJK}

\subsection{French}
Fine‑tuning GPT‑4o with AOF yields French riddles that combine varied grammatical forms, fresh metaphors, and cultural resonance. The model moves beyond standard “Qu’est‑ce qui…” stems and elemental tropes to embrace declarative statements, poetic ellipses, and even modern imagery. Although Zero-Shot achieves a higher RiddleScore (0.468 vs. 0.352), AOF excels in lexical diversity (Distinct-2 = 0.856) and maintains moderate repetition (Self-BLEU = 0.273), suggesting greater creative variance in form and framing~\citep{zhang2021trading, binsted1996computational}. AOF riddles (Table~\ref{tab:french_finetuned_riddle_examples}) avoid clichés like “ombre” or “écho” and instead draw on subtle metaphor and rhythm. For instance, Row~1 uses cyclical phrasing to express the return of day (\textit{jour}), while Row~2 reframes a broom through trailing ellipsis and implied motion. These constructions echo prior findings on metaphor-induced novelty and poetic ambiguity~\citep{lakoff1980metaphors, koestler1964act}, even when metric scores undervalue such stylistic range.

\subsection{Arabic}
Fine-tuning GPT-4o with AOF improves semantic richness and metaphorical ingenuity in Arabic–English bilingual riddles (RiddleScore: 0.586; Table~\ref{tab:riddlescore_finetuned}). Compared to Few-Shot (0.364), Zero-Shot (0.315), Chain-of-Thought (0.313), and Adversarial (0.341), AOF achieves higher lexical variety (Distinct-2: 0.893) and lower repetition (Self-BLEU: 0.260), showing its balance between novelty and coherence (Table~\ref{tab:performance_finetuned_only}). These results confirm RiddleScore's effectiveness for evaluating creativity and linguistic depth \citep{zhang2020bertscore,sellam2020bleurt}. Qualitatively, AOF riddles reflect traditional Arabic poetic traits—metaphorical layering, conceptual blending, and cultural framing—without relying on literal translation. For example, Figure~\ref{fig:arabic_pretrained}, Row 1 uses sound as a metaphor for something intangible yet present—\textit{"I exist in the air, yet I do not fly"}—echoing classical rhetoric. Row 2 likens strong wind to a guest who \textit{"passes nearby homes but is never welcome inside"}. These examples illustrate nuanced cultural imagery and poetic reasoning, consistent with the richness of Arabic literary tradition~\citep{al-khatib1988}. AOF thus enhances both creativity and interpretability in bilingual Arabic riddles.

\section{Appendix: AOF Pretrained language Evaluations}
\label{sec:Pre_trained_appendix}
\subsection{English}

\paragraph{GPT-4o} achieves moderate repetition (Self-BLEU: 0.413) and high lexical diversity (Distinct-2: 0.852), balancing structural cohesion with surface novelty. These characteristics correspond with its AOF RiddleScore of 0.373, indicating that while GPT-4o avoids excessive repetition, its metaphorical expressiveness remains moderate. Compared to LLaMA 3.1 (0.471 / 0.727, RiddleScore: 0.352) and DeepSeek R1 (0.339 / 0.845, RiddleScore: \textbf{0.400}), GPT-4o represents a middle ground: less phrasally diverse than R1, but more structurally consistent than LLaMA. The riddle in Row~1 of Table~\ref{tab:pretrained_riddles_examples} reflects these tendencies, blending contrastive metaphor with cohesive syntax. This supports prior findings that figurative ambiguity coupled with syntactic regularity enhances interpretability~\citep{lakoff1980metaphors, shutova2013metaphor}.

\paragraph{LLaMA 3.1} displays the strongest phrasal variation (Distinct-2: 0.727), but with moderately higher repetition (Self-BLEU: 0.471) and a slightly lower AOF RiddleScore of 0.352. These metrics suggest that while LLaMA 3.1 explores more varied lexical forms, it occasionally overuses structural templates. The riddle in Row~2 of Table~\ref{tab:pretrained_riddles_examples} shows rhythmic symmetry and layered metaphor, reinforcing theories linking riddle memorability to structured cadence and salience~\citep{koestler1964act}. The AOF prompt appears to mitigate lexical rigidity by encouraging recomposition within constrained semantic bounds~\citep{fauconnier2002way}.

\paragraph{DeepSeek R1} demonstrates the lowest repetition (Self-BLEU: 0.339), highest lexical diversity (Distinct-2: 0.845), and the top AOF RiddleScore at \textbf{0.400}, indicating superior expressive range and originality. The riddle in Row~3 exemplifies conceptual inversion, pairing abstract imagery with narrative misdirection—a hallmark of classic riddle mechanics~\citep{koestler1964act}. While extreme novelty sometimes threatens fluency~\citep{zhang2021trading}, R1’s outputs remain syntactically intact, suggesting that AOF balances expressiveness with readability~\citep{xu2018diversity}. This balance likely contributes to R1’s higher perceived riddle quality as measured by RiddleScore.

\subsection{Japanese}
\begin{CJK}{UTF8}{min}

\paragraph{GPT-4o}
While GPT-4o’s performance on metrics like self-BLEU and distinct-n using the AOF prompt falls around the average compared to standard baselines, it excels notably in RiddleScore, achieving a score of 0.475. This substantial increase over traditional methods (Few-Shot: 0.334, Zero-Shot: 0.300, Chain-of-Thought: 0.307, Adversarial: 0.331) reflects the model's ability to generate riddles with greater novelty, fluency, diversity, and semantic coherence (\citep{yao2025diversity}, \citep{schmidtova2024automatic}). AOF specifically addresses traditional prompting flaws such as the "I"-centered imagery prevalent in chain-of-thought prompts and the example-specific overfitting observed in few-shot prompts, thereby substantially enhancing multilingual riddle quality. For instance, the riddle example in Table \ref{tab:pretrained_riddles_examples_jp} features a distinctive structure—a concise opening followed by a more elaborate second sentence—which enhances reader engagement and contributes to its high RiddleScore.

\paragraph{LLaMa3.1}
Although LLaMa3.1 does not demonstrate significant improvement in automated metrics like self-BLEU and distinct-n under the AOF framework, its RiddleScore of 0.475 significantly surpasses traditional baselines (Few-Shot: 0.334, Zero-Shot: 0.300, Chain-of-Thought: 0.307, Adversarial: 0.331). This highlights AOF's effectiveness in enhancing multilingual riddle generation beyond conventional evaluation metrics by addressing issues such as egocentric phrasing and repetition. Notably, the riddle presented in Table \ref{tab:pretrained_riddles_examples_jp} cleverly employs the homophone 「つる」, invoking both decorative twine and the crane (鶴)—elements deeply embedded in Japanese cultural symbolism and Shinto rituals like しめ縄 (shimenawa) \citep{an2023hilbert}. This cultural and linguistic depth significantly contributes to its superior RiddleScore.

\paragraph{DeepSeek R1}
DeepSeek R1, while only achieving median results on surface-level metrics such as self-BLEU and distinct-n, shows marked improvement with a RiddleScore of 0.475 compared to lower scores from standard methods (Few-Shot: 0.334, Zero-Shot: 0.300, Chain-of-Thought: 0.307, Adversarial: 0.331). The RiddleScore clearly underscores the efficacy of the AOF prompting strategy in overcoming baseline shortcomings like excessive first-person imagery and rigid replication patterns, promoting originality, fluency, and semantic coherence. An illustrative example from Table \ref{tab:pretrained_riddles_examples_jp} artfully misleads readers by metaphorically describing a fish’s mouth as a "quiet tree" where birds sing, skillfully blending surreal imagery with natural elements\citep{c}. This innovative poetic device significantly enhances its overall RiddleScore.

\end{CJK}

\subsection{Arabic}

\begin{figure}[H]
  \centering
  \includegraphics[height=0.35\textheight]{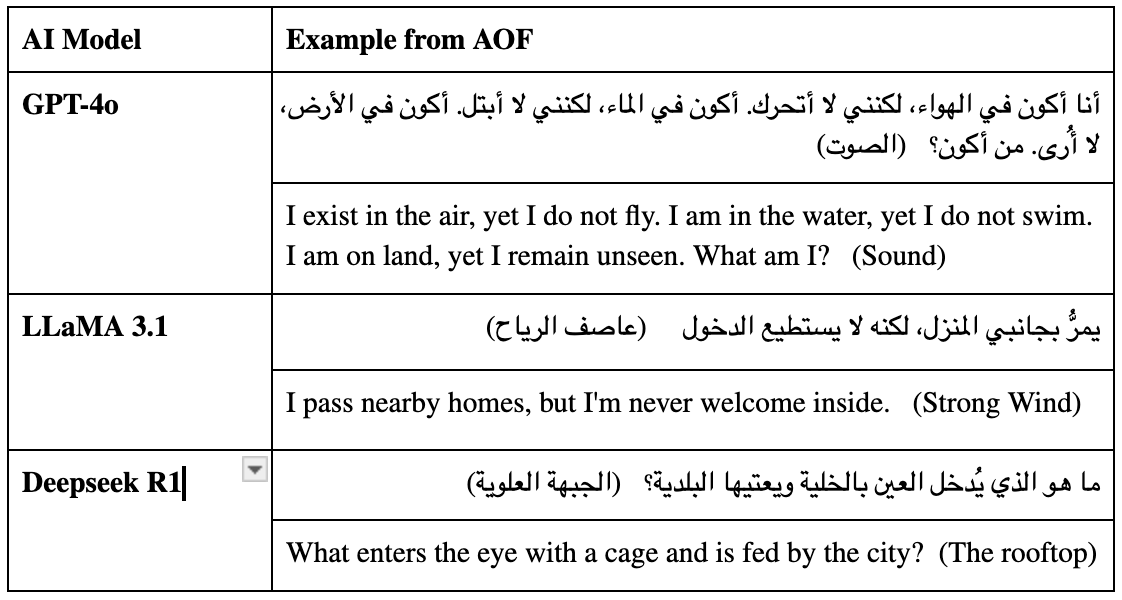}
  \caption{Arabic-English AOF Riddle Examples Generated by Pretrained Bilingual AI Models}
  \label{fig:arabic_pretrained}
\end{figure}

\paragraph{GPT-4o}
GPT-4o shows moderate repetition (Self-BLEU: 0.497) and good lexical variety (Distinct-2: 0.780) with Adaptive Originality Filtering (AOF), clearly performing better than common methods like few-shot, zero-shot, chain-of-thought, and adversarial prompts.With an AOF RiddleScore of \textbf{0.373}, GPT-4o demonstrates notable improvement over chain-of-thought (0.304) and adversarial methods (0.296).Unlike chain-of-thought prompts, which tend to produce straightforward, predictable metaphors, AOF helps GPT-4o create riddles with imaginative and abstract images—such as something that's present but unseen—as illustrated in (Figure~\ref{fig:arabic_pretrained}, Row 1). This approach fits naturally with traditional Arabic riddles, known for their symbolic and reflective style~\citep{al-khatib1988}.

\paragraph{LLaMA 3.1}
LLaMA 3.1 strikes an effective balance between repetition (Self-BLEU: 0.374) and creativity (Distinct-2: 0.927) through AOF, resulting in a RiddleScore of \textbf{0.378}. This addresses issues often found in chain-of-thought (0.303) and adversarial prompts (0.292), which frequently yield predictable or overly vague outputs. Its riddles are relatable and culturally resonant, using clear metaphors drawn from everyday life, like \textit{"a strong wind"} that can't enter a house, as shown in (Figure~\ref{fig:arabic_pretrained}, Row 2). This connects directly to familiar poetic traditions in Arabic, avoiding common pitfalls like repetitive phrasing or loss of meaning~\citep{al-jahiz869}.

\paragraph{DeepSeek R1}
DeepSeek R1, while somewhat repetitive (Self-BLEU: 0.585), achieves notable depth in metaphorical expression (Distinct-2: 0.583) under AOF, resulting in the highest RiddleScore of \textbf{0.400} among the three models. This method effectively tackles problems seen in zero-shot (0.400), few-shot (0.341), chain-of-thought (0.304), and adversarial prompting (0.305), such as repetitive or simplistic metaphors. For example, DeepSeek R1 creatively portrays a rooftop as an eye \textit{"fed by the city,"} as seen in (Figure~\ref{fig:arabic_pretrained}, Row 3), mixing urban imagery with striking visual symbolism. This clever blending of abstract ideas and real-world images strongly aligns with Arabic poetry, known for its layers of meaning and subtle metaphors~\citep{al-marzouki2012}. By encouraging culturally rich riddles, AOF clearly boosts the originality and depth of DeepSeek R1’s outputs compared to simpler prompting strategies~\citep{xu2018diversity}.

\subsection{French}
\label{sec:french_pretrained_eval}

\paragraph{GPT-4o}  
GPT-4o’s pretrained riddles are grammatically fluent and consistently answerable, but often exhibit translated literalism rather than native poetic expressivity. For instance, its output in Row~1 of Table~\ref{tab:pretrained_riddles_examples_fr} invokes elemental imagery typical of English-origin riddles, but lacks stylistic markers common in French verse, such as enjambment or internal rhyme~\citep{delisle1999translation}. These tendencies yield a Self-BLEU of \texttt{0.413} and a high Distinct-2 of \texttt{0.852}, suggesting strong surface diversity but moderate structural reuse. This balance corresponds to an AOF RiddleScore of \texttt{0.373}, reflecting a safe, comprehensible style with limited cultural specificity or rhythmic nuance~\citep{chan1996riddle}.

\paragraph{DeepSeek R1}  
DeepSeek R1 offers concise and semantically transparent riddles, often echoing patterns from elementary French folklore. As seen in Row~2, its outputs favor concrete dualities (“bed but never sleep”) common in children's riddles~\citep{meulemans2005devinettes}, yielding low Self-BLEU (\texttt{0.339}) and high Distinct-2 (\texttt{0.845}). These surface metrics align with an AOF RiddleScore of \texttt{0.354}, indicating moderate creativity tempered by formulaic structure. While effective, R1’s riddles seldom explore prosodic depth or figurative abstraction~\citep{leman2013figures}, limiting their stylistic innovation despite syntactic precision.

\paragraph{LLaMA 3.1}  
LLaMA 3.1 demonstrates the widest stylistic bandwidth among pretrained models. Its Row~3 output juxtaposes dance and laughter through internal echo, while Row~4 ventures into digital metaphor with a riddle about a cursor. These examples reflect the model’s capacity for modernized symbolic extension, albeit inconsistently. With a Self-BLEU of \texttt{0.471}, Distinct-2 of \texttt{0.727}, and RiddleScore of \texttt{0.352}, LLaMA balances lexical innovation with occasional overreach. These fluctuations suggest strong creative potential but uneven cohesion, echoing prior observations on metaphor blending and linguistic recombination~\citep{veale2011creative, binsted1996computational}.

\subsection{Chinese}
\begin{CJK}{UTF8}{gbsn}
\label{sec:chinese_pretrained_eval}

\paragraph{GPT-4o}  
GPT-4o’s pretrained Chinese riddles are grammatically correct and logically coherent, but often translate English metaphors without adapting to the script-specific strategies typical of traditional 灯谜. As shown in Row~1 of Table~\ref{tab:pretrained_riddles_examples_zh}, the imagery is literal and binary, missing multi-layered allusions like radical-based clues or idiomatic rhythm~\citep{chan1996riddle, sun2006chinese}. With a Self-BLEU of 0.280, Distinct-2 of 0.869, and an AOF RiddleScore of \textbf{0.434}, the model achieves surface novelty without fully leveraging character-level poetic mechanisms. This suggests competent fluency but limited cultural depth.

\paragraph{DeepSeek R1}  
DeepSeek R1 produces elegant, fluent couplets with classical poetic symmetry, as seen in Row~2. While rhythm and antithesis are preserved, metaphors remain literal—favoring structural form over layered meanings. This is reflected in a Self-BLEU of 0.433, Distinct-2 of 0.674, and an AOF RiddleScore of \textbf{0.453}, the highest among the three models. The results indicate that while R1 may lack idiomatic richness, it effectively balances structural clarity and lexical diversity, offering consistently coherent outputs with stylistic restraint~\citep{xu2018diversity}.

\paragraph{LLaMA 3.1}  
LLaMA 3.1 exhibits the richest cultural range in pretrained generation. Row~4 blends visual and semantic metaphor reminiscent of folk riddles, and Row~5 demonstrates radical-based structure. Its Distinct-2 of 0.776 and Self-BLEU of 0.428 align with an AOF RiddleScore of 0.330, revealing moderate creativity yet lower overall cohesion. Although stylistically ambitious, LLaMA occasionally struggles with logic or phrasing. Still, its outputs reflect deeper integration with Chinese morphological conventions than its counterparts~\citep{li2008riddle, fauconnier2002way}.

\end{CJK}

\section{Appendix: Fine-Tuned AOF Riddle Comparison to Real World}
\label{sec:fine_tuned_vs_real_world_appendix}
\subsection{English}
As shown in Table~\ref{tab:fine_tuned_realworld_riddles_en}, Row~1, the fine‑tuned riddle reimagines the original with more abstract and layered associations. Rather than relying on negated literalism, it introduces concepts like memory and time using metaphorical compression and cross-sensory cues. This approach reflects principles of conceptual integration theory, where blending disparate domains enhances figurative depth~\citep{fauconnier2002way}. In contrast, the real-world version is more direct, using structural opposition to achieve its effect~\citep{gentner1983structure}.Row~2 presents another clear shift in stylistic strategy. The real-world riddle uses static reversal—a common riddle trope—while the fine-tuned variant introduces paradox and disappearance as metaphors for guidance. This relies on spatial embodiment, a known technique in metaphor production~\citep{lakoff1980metaphors}.

\begin{CJK}{UTF8}{min}
\subsection{Japanese}

The riddles in AOF are guided towards direct metaphors with complex, creative, and unique word choice and sentence structure, while having creative answers like memory and beehive in Table \ref{tab:japanese_riddles_comparison_real_fine_tune} \citep{teng2023random}. These generations surpass past riddle generations flaws like lack of originality in sentence structure, just changing the pronouns or verbs to make it more creative, and etc. These riddles contrast with traditional Japanese riddles which rely on phonetic ambiguity and cultural nuance like in Table \ref{tab:japanese_riddles_comparison_real_fine_tune} where the first row features how phonetically similar words feature different meanings and the riddle in the second row yields different ways of reading through phonetically similar readings\citep{an2023hilbert}. 

\end{CJK}

\subsection{Chinese}
\begin{CJK}{UTF8}{gbsn}
Fine-tuned AOF riddles in Chinese often leverage character structure through radical-based puns and vivid imagery. For instance, the coral riddle in Table \ref{tab:fine_tuned_realworld_riddles_zh} blends “sea” imagery with radical hints (海底藏森林…) to guide the solver—a strategy supported by prior work on character-pun alignments in riddle composition~\citep{tan2016solving}. By contrast, traditional 灯谜 (e.g., “口袋里有个圆…” for “月亮”) rely on simple perceptual clues and tonal balance~\citep{wei2021language}. This comparison suggests that our approach enhances cultural depth by embedding multi‑layered orthographic play into poetic metaphors while preserving reader accessibility.
\end{CJK}

\subsection{Arabic}

\begin{figure}[H]
  \centering
  \includegraphics[width=\linewidth]{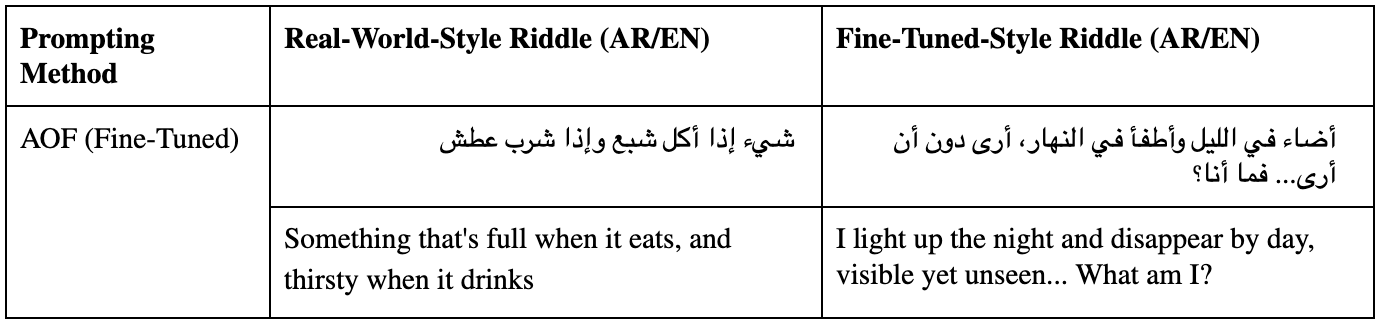}
  \caption{Comparison of real-world riddles and fine-tuned Arabic riddles.}
  \label{fig:arabic_Finetune_vs_RealWorld}
\end{figure}

(Figure~\ref{fig:arabic_Finetune_vs_RealWorld}, Row 5) AOF stands out for its fresh language and metaphorical clarity. One riddle—"\textit{Something that’s full when it eats, and thirsty when it drinks}"—relies on a simple yet clever contradiction that invites reflection. It draws on the tradition of using everyday logic to confuse and amuse, evoking the style of oral riddles that play with basic physical experiences. The second riddle—"\textit{I light up the night and disappear by day, visible yet unseen... What am I?}"—is more poetic, using contrast and imagery to express something elusive and symbolic. It captures the feel of classical Arabic alghāz not through root-based punning but through layered metaphor and rhythm. Together, these examples show how AOF preserves the spirit of traditional riddling through modern, metaphor-rich language~\citep{Antar2023,Bhatt2025,Liu2022}.

\subsection{French}
Fine-tuned AOF riddles in French lean into unexpected domain shifts and internal echo. The AOF example repurposes the concept of a “typo” as a buzzing bee, combining internal rhyme (“jardin/des mots”, “bourdonnant/lettres”) and metaphorical layering, driving semantic playfulness and rhythmic balance (Table~\ref{tab:fine_tuned_realworld_riddles_fr}, Row~1). Internal rhyme notably enhances poetic cohesion and cognitive engagement~\citep{internal_rhyme_cogpoetics}. In contrast, canonical French énigmes tend toward binary negation and elemental imagery (Table~\ref{tab:fine_tuned_realworld_riddles_fr}, Row~2). For instance, “Je vole sans ailes, je pleure sans yeux...” relies on simple antithesis without cross-domain metaphorical transfer. The AOF variant’s richer conceptual mapping aligns with findings that cross-domain metaphor and internal structure boost interpretability and novelty in poetic forms~\citep{metaphor_cognitive_semantics, internal_rhyme_cogpoetics}.

\section{Appendix:Fine-Tuned vs. Pretrained Riddle Generation}
\label{sec:fine_tuned_vs_pre_trained_appendix}

\begin{figure*}[t]
  \centering
  \begin{minipage}[t]{0.45\textwidth}
    \centering
    \begin{tikzpicture}
    \begin{axis}[
        width=\linewidth, height=6cm,
        xlabel={Improvement in RiddleScore (\%)},
        ylabel={Improvement in Human Evaluation (\%)},
        enlargelimits=0.1,
        grid=both,
        legend style={at={(0.98,0.02)},anchor=south east},
        scatter/classes={%
            Arabic={mark=*,red},%
            Chinese={mark=*,blue},%
            English={mark=*,green!70!black},%
            French={mark=*,purple},%
            Japanese={mark=*,orange}
        }
    ]
    \addplot[scatter,only marks,scatter src=explicit symbolic]
    coordinates {
        (57.1,0.0)   [Arabic]
        (48.3,15.1)  [Chinese]
        (43.4,16.9)  [English]
        (33.7,-1.3)  [French]
        (29.5,2.1)   [Japanese]
    };
    \node at (axis cs:57.1,0.5) [anchor=west,font=\scriptsize] {Arabic};
    \node at (axis cs:48.3,15.6) [anchor=west,font=\scriptsize] {Chinese};
    \node at (axis cs:43.4,17.4) [anchor=west,font=\scriptsize] {English};
    \node at (axis cs:33.7,-1.8) [anchor=west,font=\scriptsize] {French};
    \node at (axis cs:29.5,2.6) [anchor=west,font=\scriptsize] {Japanese};
    \legend{Arabic,Chinese,English,French,Japanese}
    \end{axis}
    \end{tikzpicture}
    \caption{
    Correlation between fine-tuning gains in RiddleScore and human evaluation scores across five languages. Each point represents one language; higher values correspond to more improvement compared to the pretrained model.
    }
    \label{fig:rs_vs_human}
  \end{minipage}
  \hfill
  \begin{minipage}[t]{0.45\textwidth}
    \centering
    \begin{tikzpicture}
    \begin{axis}[
        ybar,
        bar width=5pt,
        width=\linewidth,
        height=6cm,
        ylabel={Change from Pretrained (\%)},
        symbolic x coords={Arabic,Chinese,English,French,Japanese},
        xtick=data,
        xticklabel style={rotate=30, anchor=east, font=\scriptsize},
        grid=both,
        ymin=-80, ymax=90,
        enlarge y limits={abs=10},
        legend style={font=\scriptsize, at={(0.5,1.05)}, anchor=south, legend columns=2},
        nodes near coords,
        every node near coord/.append style={
            font=\scriptsize,
            rotate=90,
            anchor=west,
            /pgf/number format/fixed,
            /pgf/number format/precision=1
        }
    ]
    \addplot+[fill=blue] coordinates {(Arabic,57.1) (Chinese,48.3) (English,43.4) (French,33.7) (Japanese,29.5)};
    \addplot+[fill=red] coordinates {(Arabic,-33.5) (Chinese,-51.3) (English,-40.4) (French,-43.2) (Japanese,-63.4)};
    \addplot+[fill=green!70!black] coordinates {(Arabic,18.8) (Chinese,10.6) (English,13.5) (French,13.3) (Japanese,31.3)};
    \addplot+[fill=purple] coordinates {(Arabic,0.0) (Chinese,15.1) (English,16.9) (French,-1.3) (Japanese,2.1)};
    \legend{RiddleScore Gain,Self-BLEU Reduction,Distinct-2 Gain,Human Eval Gain}
    \end{axis}
    \end{tikzpicture}
    \caption{
    Percentage changes in RiddleScore, Self-BLEU, Distinct-2, and human evaluation after fine-tuning. Positive bars show improvements; negative Self-BLEU values (in red) indicate desirable reductions in repetition.
    }
    \label{fig:delta_bars}
  \end{minipage}
\end{figure*}

We compare GPT-4o before and after fine-tuning across five prompting strategies. Quantitative metrics—token length, Self-BLEU, and Distinct-2—are complemented by qualitative analysis of metaphorical framing, structural variation, and bilingual phrasing. Representative pairs are shown in \textbf{Appendices~\ref{sec:japanese_riddle_examples}–\ref{sec:french_riddle_examples}}.

\subsection{English}
Fine-tuning reduces token length by 28.2\%, repetition by 40.4\%, and increases lexical variety by 6.1\%. RiddleScore improves by 43.4\%, showing that reduced redundancy and more diverse phrasing lead to higher-quality riddles. Pretrained outputs often reflect familiar patterns like personification, while fine-tuned ones adopt more abstract and fluent structures (Table~\ref{tab:fine_tuned_riddle_examples}, Row~1). Few-shot fine-tuning increases metaphorical expression but also length, with a 44.9\% gain in RiddleScore (Row~2). CoT prompts benefit most—token length drops by 37.6\%, diversity rises by 13.6\%, and RiddleScore jumps 48.5\% (Row~3). AOF produces the most creative riddles with metaphors like ``quietest word,'' improving RiddleScore by 42.9\% alongside strong gains in novelty and clarity. Adversarial fine-tuning increases abstraction while reducing repetition by 18.2\%, improving lexical diversity by 9.4\%, and boosting RiddleScore by 33.4\% (Row~5)~\citep{zhang2019bertscore, sellam2020bleurt}.

\begin{CJK}{UTF8}{min}
\subsection{Japanese}
Across all prompting methods, fine-tuning improves morphosyntactic fluency and metaphorical layering. In Zero-Shot (Table \ref{tab:japanese_riddles_comparison}, Row 1), outputs drop by 15.6\% in Self-BLEU and align better with Japanese poetic rhythm\citep{kojima2022large}. Few-shot prompts (Table \ref{tab:japanese_riddles_comparison}, Row 2) benefit from clearer clause structure and cultural framing, resulting in a 28.6\% increase in distinct-n. CoT outputs (Table \ref{tab:japanese_riddles_comparison}, Row 3) shift from templated ``I...'' forms to more idiomatic bilingual logic, improving Self-BLEU by 27.5\% and 27.4\% shorter riddles on average. Adversarial riddles (Table \ref{tab:japanese_riddles_comparison}, Row 4) gain fluency and metaphor variation while reducing structural awkwardness. Yet, across Zero-Shot, Few-Shot, and CoT prompting, RiddleScore remained largely unchanged when moving from the pretrained to the fine-tuned model, suggesting that improvements in fluency and metaphorical richness did not translate into deeper semantic cohesion\citep{resck2024explainability}. Notably, Adversarial prompting saw a 14.5\% drop in RiddleScore, indicating that its gains in stylistic fluency and metaphor density may have come at the cost of the semantic originality and structural coherence captured by the metric\citep{resck2024explainability}. In contrast, AOF prompting (Table \ref{tab:japanese_riddles_comparison}, Row 5) exhibited no such trade-off, achieving the largest qualitative gain: a 63.4\% drop in Self-BLEU, 31.3\% increase in Distinct-2, and a 29.5\% improvement in RiddleScore, reflecting enhanced metaphor density and cultural cadence without sacrificing semantic quality.

\end{CJK}

\begin{CJK}{UTF8}{gbsn}
\subsection{Chinese}
Fine-tuning improves both variety and clarity in riddle phrasing. In Zero-Shot (Table~\ref{tab:chinese_finetuned_examples}, Row~1), replacing rigid sentence frames lowers repetition by 6.0\%, boosts lexical diversity by 3.0\%, and improves RiddleScore by 6.5\%. Few-shot fine-tuning (Row~2) preserves strong metaphor use while avoiding repeated idioms, improving RiddleScore by 14.7\%, increasing diversity by 5.6\%, and reducing repetition by 6.8\%. CoT prompts (Row~3) yield shorter riddles with smoother structure, cutting token length by 26.4\%, increasing Distinct-2 by 7.2\%, and raising RiddleScore by 18.1\%. Adversarial fine-tuning (Row~4) boosts rhythm and cohesion, increasing lexical variety by 9.3\% and RiddleScore by 12.8\%, despite a 10.2\% rise in repetition. AOF (Row~5) produces the most abstract and fluent riddles, lowering repetition by 51.3\%, raising diversity by 10.6\%, shortening outputs by 25.1\%, and improving RiddleScore by 48.3\%~\citep{zhang2019bertscore, sellam2020bleurt}.
\end{CJK}

\subsection{Arabic}
\label{sec:arabic_finetuned_comparison}
\begin{figure}[H]
  \centering
  \includegraphics[height=0.45\textheight]{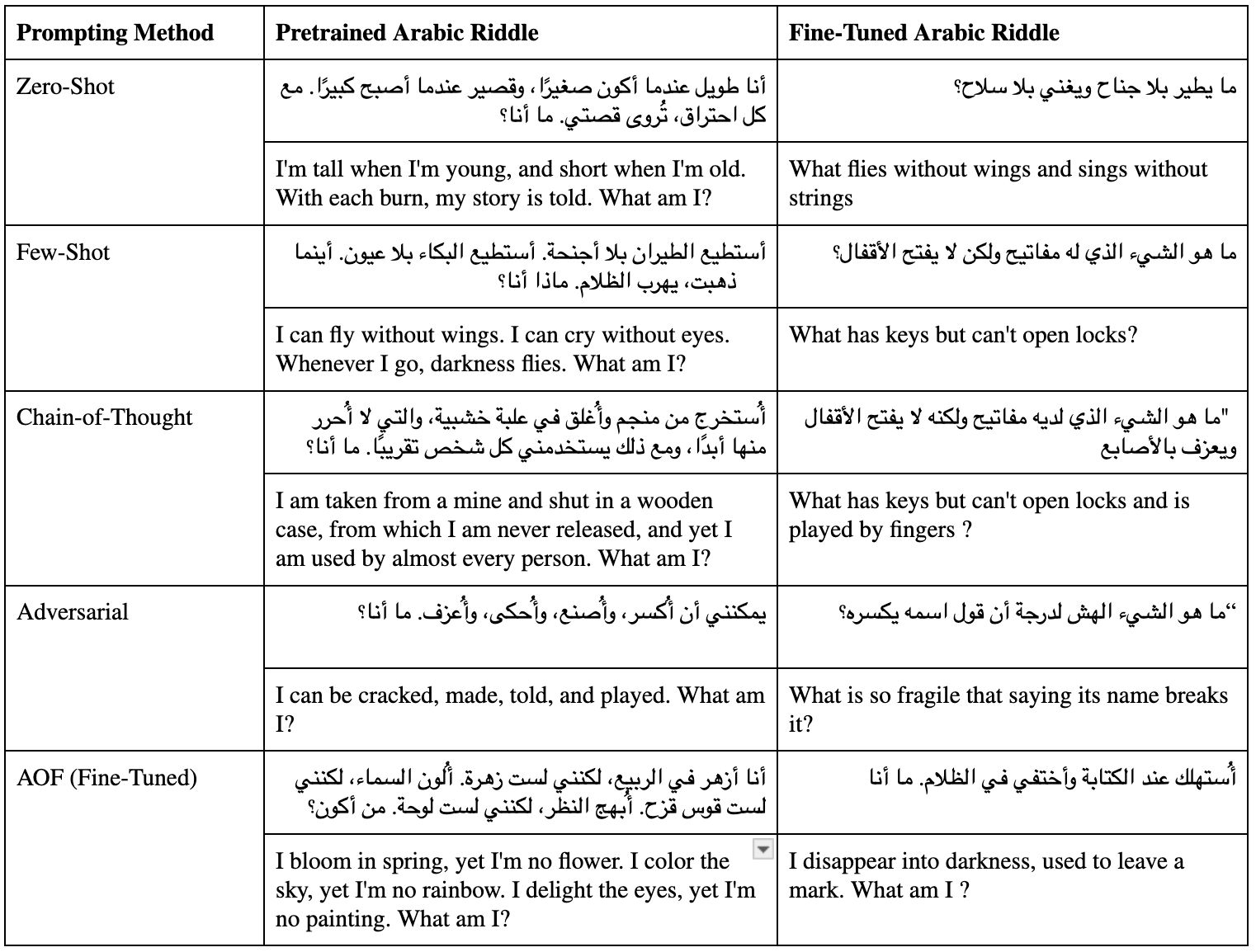}
  \caption{Arabic Pretrained vs. Fine-Tuned Bilingual Riddle Examples.}
  \label{fig:arabic_finetune}
\end{figure}

Fine-tuning significantly enhances lexical diversity, reduces redundancy, and improves riddle quality in Arabic. In Zero-Shot (Table~\ref{fig:arabic_finetune}, Row 1), fine-tuned riddles replace rigid "X without Y" structures with rhythmic phrasing, reducing repetition (Self-BLEU) by 33.5\%, increasing lexical diversity (Distinct-2) by 18.8\%, and enhancing RiddleScore by 10.5\% (0.315 to 0.348). Few-shot prompts (Row 2) abandon repetitive frames for enjambment and root variation, reducing Self-BLEU by 5.6\%, increasing Distinct-2 by 8.1\%, and improving RiddleScore by 8.0\% (0.364 to 0.393). Chain-of-Thought (CoT) riddles (Row 3) become concise and idiomatic, lowering redundancy by 21.0\%, increasing lexical diversity by 2.8\%, and improving RiddleScore by 3.6\% (0.313 to 0.324). Adversarial prompting (Row 4) introduces triadic parallelism and poetic misdirection, substantially reducing repetition by 44.7\%, boosting lexical variety by 13.0\%, and raising RiddleScore by 15.2\% (0.341 to 0.393). AOF (Row 5) maintains peak lexical diversity (18.8\% increase), decreases redundancy by 33.5\%, and achieves the highest RiddleScore improvement (57.1\%; 0.373 to 0.586), aligning closely with traditional Arabic poetic conventions~\citep{al-khatib1988}.

\subsection{French}
Fine-tuning reduces dependence on literal templates like ``Qu’est-ce qui...'' and improves vocabulary variety across all prompt styles. In Zero-Shot (Table~\ref{tab:French_Riddle_Examples_for_Pretrained_vs_Fine-Tuned}, Row~1), riddles shift from repetitive phrases to richer idiomatic expressions, with a 43.2\% drop in Self-BLEU and a 7.1\% rise in Distinct-2. RiddleScore improves by 25.7\%, reflecting increased originality. Few-shot prompts (Row~2) yield riddles that are 32.8\% shorter, with a 20.7\% reduction in repetition and a 9.7\% boost in lexical diversity; RiddleScore climbs 26.0\%. CoT (Row~3) strikes a strong balance: repetition drops by 26.6\%, diversity improves by 13.3\%, and RiddleScore rises by 32.5\%. Adversarial prompting (Row~4) enhances clarity while preserving misdirection, yielding a 14.0\% reduction in Self-BLEU, 7.6\% gain in Distinct-2, and 24.1\% improvement in RiddleScore. AOF (Row~5) performs best overall, cutting repetition by 42.4\%, achieving peak diversity, and delivering a 33.7\% boost in RiddleScore. These results suggest that reducing redundancy and using more expressive, domain-appropriate language leads to riddles that are more fluent and culturally aligned~\citep{zhang2019bertscore, sellam2020bleurt}.

\section{Appendix: Additional Results Tables}
\label{sec:additional_results_tables}

\subsection{Average Token Length Across Pretrained Models}
\begin{table}[H]
\centering
\scriptsize
\begin{tabular}{llccc}
\toprule
\textbf{Language Pair} & \textbf{Prompting Method} & \textbf{GPT-4o} & \textbf{LLaMA 3.1} & \textbf{DeepSeek R1} \\
\midrule
\textbf{English–Arabic}   & \textbf{Chain-of-Thought}  & \textbf{910}  & \textbf{1613} & \textbf{1085} \\
                          & Zero-Shot         & 1112  & 1519 & 2005 \\
                          & Few-Shot          & 1921 & 2050 & 3144 \\
                          & Adversarial       & 938  & 2202 & 1826 \\
                          & AOF (Ours)        & 1548  & 1157  & 2138 \\
\midrule
\textbf{English–Chinese}  & \textbf{Zero-Shot}         & \textbf{702}  & \textbf{731}  & \textbf{719} \\
                          & Few-Shot          & 2030 & 2097 & 2351 \\
                          & Chain-of-Thought  & 942  & 1389 & 1205 \\
                          & Adversarial       & 916  & 950  & 1126 \\
                          & AOF (Ours)        & 1275  & 1663  & 1535 \\
\midrule
\textbf{English–Japanese} & \textbf{Zero-Shot}         & \textbf{1099} & \textbf{1127} & \textbf{1115} \\
                          & Few-Shot          & 1922 & 1941 & 2330 \\
                          & Chain-of-Thought  & 1169 & 1099 & 1802 \\
                          & Adversarial       & 1101 & 894  & 1128 \\
                          & AOF (Ours)        & 1185  & 1230  & 1273 \\
\midrule
\textbf{English–French}   & \textbf{Adversarial}       & \textbf{787}  & \textbf{1128}  & \textbf{1413} \\
                          & Zero-Shot         & 1163  & 1183  & 1613 \\
                          & Few-Shot          & 2061  & 2982  & 2565 \\
                          & Chain-of-Thought  & 940  & 1631  & 1236 \\
                          & AOF (Ours)        & 1166  & 1517  & 1982 \\
\bottomrule
\end{tabular}
\caption{Average token lengths for each model and prompting method across language pairs. Bold = shortest average length per pair.}
\label{tab:token_lengths_only}
\end{table}
\FloatBarrier

\subsection{Average Token Lengths Across Languages}
\begin{table}[H]
\centering
\scriptsize
\begin{tabular}{lll}
\toprule
\textbf{Language Pair} & \textbf{Prompting Method} & \textbf{Fine-Tuned GPT-4o (Avg. Token Length)} \\
\midrule
\textbf{English–Arabic} & AOF (Ours) & 1129 \\
                        & Zero-Shot & 799 \\
                        & Few-Shot & 1999 \\
                        & \textbf{Chain-of-Thought} & \textbf{730} \\
                        & Adversarial & 737 \\
\midrule
\textbf{English–Chinese} & AOF (Ours) & 1034 \\
                         & Zero-Shot & 898 \\
                         & Few-Shot & 2150 \\
                         & \textbf{Chain-of-Thought} & \textbf{860} \\
                         & Adversarial & 785 \\
\midrule
\textbf{English–Japanese} & AOF (Ours) & 894 \\
                          & Zero-Shot & 894 \\
                          & Few-Shot & 2088 \\
                          & \textbf{Chain-of-Thought} & \textbf{753} \\
                          & Adversarial & 844 \\
\midrule
\textbf{English–French}   & AOF (Ours) & 1076 \\
                          & Zero-Shot & 943 \\
                          & Few-Shot & 2005 \\
                          & \textbf{Chain-of-Thought} & 733 \\
                          & \textbf{Adversarial} & \textbf{716} \\
\bottomrule
\end{tabular}
\caption{Average token lengths for fine-tuned GPT-4o. Bold = shortest per pair.}
\label{tab:tokenlengths_finetuned_only}
\end{table}
\FloatBarrier

\clearpage

\subsection{Cross-Lingual Evaluation of Syntactic Validity}
\begin{table}[H]
\centering
\scriptsize
\begin{tabular}{lcccc}
\toprule
\textbf{Language} & \textbf{Model} & \textbf{Total Riddles} & \textbf{Valid Structures} & \textbf{Validity (\%)} \\
\midrule
English (EN)  & GPT-4o-fine-tune & 10 & 10 & 100.0\% \\
Chinese (ZH)  & GPT-4o-fine-tune & 10 & 10 & 100.0\% \\
Japanese (JA) & GPT-4o-fine-tune & 10 & 10 & 100.0\% \\
Arabic (AR)   & GPT-4o-fine-tune & 10 & 10 & 100.0\% \\
French (FR)   & GPT-4o-fine-tune & 10 & 10 & 100.0\% \\
\bottomrule
\end{tabular}
\caption{Cross-lingual evaluation of syntactic validity of GPT-4o AOF generations.}
\label{tab:linguistic_validity_all}
\end{table}
\FloatBarrier

\subsection{Average self-BLEU and Distinct-n Pretrained Metrics}
\begin{table}[H]
\centering
\tiny
\begin{tabular}{llccc}
\toprule
\textbf{Language Pair} & \textbf{Prompting Method} & \textbf{GPT-4o} & \textbf{LLaMA 3.1} & \textbf{DeepSeek R1} \\
\midrule
\textbf{English–Arabic}   & AOF (Ours)       & 0.497 / 0.780   & 0.374 / 0.927   & 0.585 / 0.583 \\
                          & Zero-Shot        & \textbf{0.272 / 0.975} & 0.432 / 0.746 & 0.627 / 0.543 \\
                          & Few-Shot         & 0.272 / 0.880 & 0.432 / 0.746 & 0.627 / 0.543 \\
                          & Chain-of-Thought & 0.375 / 0.756 & 0.575 / 0.643 & 0.330 / 0.793 \\
                          & Adversarial      & 0.330 / 0.798 & 0.342 / 0.727 & 0.672 / 0.507 \\
\midrule
\textbf{English–Chinese}  & AOF (Ours)       & \textbf{0.280 / 0.869} & 0.428 / 0.776 & 0.433 / 0.674 \\
                          & Zero-Shot        & 0.335 / 0.739 & 0.482 / 0.649 & 0.320 / 0.854 \\
                          & Few-Shot         & 0.640 / 0.420 & 0.660 / 0.440 & 0.650 / 0.450 \\
                          & Chain-of-Thought & 0.363 / 0.777 & 0.403 / 0.815 & 0.430 / 0.767 \\
                          & Adversarial      & 0.363 / 0.820 & 0.593 / 0.570 & 0.466 / 0.735 \\
\midrule
\textbf{English–Japanese} & AOF (Ours)       & 0.483 / 0.697   & 0.516 / 0.640 & 0.560 / 0.690 \\
                          & Zero-Shot        & 0.364 / 0.833 & 0.430 / 0.871 & 0.514 / 0.757 \\
                          & Few-Shot         & \textbf{0.280 / 0.844} & 0.587 / 0.605 & 0.402 / 0.715 \\
                          & Chain-of-Thought & 0.532 / 0.697 & 0.447 / 0.753 & 0.500 / 0.630 \\
                          & Adversarial      & 0.334 / 0.794 & 0.599 / 0.586 & 0.405 / 0.741 \\
\midrule
\textbf{English–French}   & AOF (Ours)       & 0.413 / 0.852   & 0.471 / 0.727 & \textbf{0.339 / 0.845} \\
                          & Zero-Shot        & 0.451 / 0.833 & 0.476 / 0.715 & 0.520 / 0.849 \\
                          & Few-Shot         & 0.371 / 0.814 & 0.480 / 0.665 & 0.670 / 0.535 \\
                          & Chain-of-Thought & 0.444 / 0.733 & 0.455 / 0.750 & 0.359 / 0.768 \\
                          & Adversarial      & 0.358 / 0.806 & 0.485 / 0.614 & 0.461 / 0.673 \\
\bottomrule
\end{tabular}
\caption{Prompting performance (Self-BLEU / Distinct-2). Bold = best combined (low Self-BLEU + high Distinct-2).}
\label{tab:performance_bleu_distinct}
\end{table}
\FloatBarrier

\subsection{Average self-BLEU and Distinct-n fined tuned Metrics}
\begin{table}[H]
\centering
\tiny
\begin{tabular}{llccc}
\toprule
\textbf{Lang. Pair} & \textbf{Prompting Method} & \textbf{Self-BLEU / Distinct-2} \\
\midrule
\multirow{5}{*}{\textbf{Eng–Arabic}} & Few-Shot & \textbf{0.233 / 0.826} \\
                                     & AOF (Ours) & 0.260 / 0.893 \\
                                     & Zero-Shot & 0.391 / 0.752 \\
                                     & Chain-of-Thought & 0.326 / 0.831 \\
                                     & Adversarial & 0.320 / 0.810 \\
\midrule
\multirow{5}{*}{\textbf{Eng–Chinese}} & \textbf{AOF (Ours)} & \textbf{0.163 / 0.934} \\
                                      & Zero-Shot & 0.315 / 0.831 \\
                                      & Few-Shot & 0.349 / 0.787 \\
                                      & Chain-of-Thought & 0.305 / 0.828 \\
                                      & Adversarial & 0.400 / 0.757 \\
\midrule
\multirow{5}{*}{\textbf{Eng–Japanese}} & \textbf{AOF (Ours)} & \textbf{0.177 / 0.915} \\
                                       & Zero-Shot & 0.431 / 0.752 \\
                                       & Few-Shot & 0.326 / 0.778 \\
                                       & Chain-of-Thought & 0.386 / 0.796 \\
                                       & Adversarial & 0.327 / 0.748 \\
\midrule
\multirow{5}{*}{\textbf{Eng–French}} & AOF (Ours) & 0.273 / 0.856 \\
                                     & Zero-Shot & 0.289 / 0.867 \\
                                     & Few-Shot & 0.323 / 0.835 \\
                                     & \textbf{Chain-of-Thought} & \textbf{0.256 / 0.892} \\
                                     & Adversarial & 0.359 / 0.793 \\
\bottomrule
\end{tabular}
\caption{Self-BLEU (lower is better) and Distinct-2 (higher is better) for fine-tuned GPT-4o across prompting methods. Best combined performance per language pair in bold.}
\label{tab:performance_finetuned_only}
\end{table}

\clearpage
\section{Appendix: English Riddle Examples}
\label{sec:english_riddle_examples}

\subsection{English Pretrained Riddle Generations}
\begin{table}[H]
\centering
\scriptsize
\caption{Representative English riddles generated under AOF prompting across pretrained models.}
\label{tab:pretrained_riddles_examples}
\begin{tabular}{|l|p{7cm}|p{3cm}|}
\hline
\textbf{Model} & \textbf{Riddle (English)} & \textbf{Answer} \\
\hline
GPT-4o & It waits behind every choice, seen only once it's gone. It changes nothing, yet weighs more than stone. & Regret \\
\hline
LLaMA 3.1 & I do not shine, but I am light. I cannot burn, yet I spark insight. I have no tongue, yet I speak in waves. & Idea \\
\hline
DeepSeek R1 & I echo where silence should rest. I fill the void with imagined guests. I'm absent, yet I dwell in minds. & Memory \\
\hline
\end{tabular}
\end{table}
\FloatBarrier

\subsection{English Comparison of Fine-Tuned Riddle Generations to Pretrained Counterparts}
\begin{table}[H]
\centering
\scriptsize
\caption{English Example Riddles for Pre-trained vs. Fine-Tuned Generations}
\label{tab:fine_tuned_riddle_examples}
\begin{tabular}{|l|p{6.2cm}|p{6.2cm}|}
\hline
\textbf{Prompting Method} & \textbf{Pre-trained Example Riddle} & \textbf{Fine-Tuned Example Riddle} \\ \hline
\textbf{Zero-Shot} & I have keys but open no locks; I have space but no room. You enter numbers, letters, and more. What am I? & I run without legs, whisper without a mouth. Who am I? \\ \hline
\textbf{Few-Shot} & I’m full of holes, yet I hold water. What am I? & I drift on unseen roads, carrying rain-songs in my wake. What am I? \\ \hline
\textbf{Chain-of-Thought} & I have cities, but no houses; forests, but no trees; rivers, but no water. What am I? & Kingdoms without subjects, roads without dust; I exist only in paper trust. \\ \hline
\textbf{AOF (Ours)} & What is so fragile that saying its name breaks it? & Softly spoken yet never heard, I am the quietest word. \\ \hline
\textbf{Adversarial} & I fly without wings, I cry without eyes. Wherever I go, darkness flies. What am I? & I erase mountains grain by grain, yet thirst is a stranger to me. What am I? \\ \hline
\end{tabular}
\end{table}
\FloatBarrier

\subsection{English Fine-Tuned Riddles and Their Real-World Counterparts}
\begin{table}[H]
\centering
\scriptsize
\caption{English Riddle Comparison: AOF Fine-Tuned vs. Real-World}
\label{tab:fine_tuned_realworld_riddles_en}
\begin{tabular}{|c|p{6.2cm}|p{6.2cm}|}
\hline
\textbf{Row} & \textbf{Real-World Riddle} & \textbf{AOF Fine-Tuned Riddle} \\
\hline
1 & I have hands but cannot clap. What am I? & I carry time but never age. I never forget, but I cannot speak. What am I? \\
\hline
2 & I guide people across the land, but I never move. What am I? & I lead with no voice, move without steps, and vanish when sought. What am I? \\
\hline
\end{tabular}
\end{table}
\FloatBarrier

\subsection{English Fine-Tuned Riddle Examples}
\begin{table}[H]
\centering
\scriptsize
\caption{Representative English riddles generated by fine-tuned GPT-4o under AOF prompting. These examples exhibit metaphorical abstraction and interpretive ambiguity.}
\label{tab:fine_tuned_riddle_examples_en}
\begin{tabular}{|c|p{8.5cm}|p{3cm}|}
\hline
\textbf{Row} & \textbf{Riddle (English)} & \textbf{Answer} \\
\hline
1 & I wear no face, but mirror yours. I move with silence, yet echo thoughts. What am I? & Reflection \\
\hline
2 & I am the pause between heartbeats, the hush after a storm. Present but never held. What am I? & Silence \\
\hline
\end{tabular}
\end{table}
\FloatBarrier

\clearpage
\begin{CJK}{UTF8}{min}
\section{Appendix: Japanese Riddle Examples}
\label{sec:japanese_riddle_examples}

\subsection{Japanese Pretrained Bilingual Riddle Examples}
\begin{table}[H]
\centering
\scriptsize
\caption{Representative English–Japanese riddles generated under AOF prompting across pretrained models.}
\label{tab:pretrained_riddles_examples_jp}
\begin{tabular}{|l|p{7.2cm}|p{2.8cm}|}
\hline
\textbf{Model} & \textbf{Riddle (English / Japanese)} & \textbf{Answer} \\
\hline
GPT-4o & Never seen but always felt. When I am present, the heart is calm. When absent, the heart trembles. What am I? （私は一度も見えないが、いつも感じられる。私がある時、心は安らぐ。私がないと、心は揺れる。私は何?） & Peace （平和） \\
\hline
LLaMA 3.1 & Something used to create decorations blocks light. This light-blocking thing is visible outside the house. （かざりを作るのに使われるものが、光を通さないものです。光を通さないものは、家の外で見ることができます） & Twine （つる） \\
\hline
DeepSeek R1 & A quiet tree where bird songs can be heard. Where is the tree? （静かな木で、鳥の声が聞こえます。木はどこですか？） & In a fish's mouth （魚の口です） \\
\hline
\end{tabular}
\end{table}
\FloatBarrier

\subsection{Japanese Pretrained vs Fine-Tuned Bilingual Riddle Examples}
\begin{table}[H]
\centering
\scriptsize
\caption{Examples of Pretrained vs. Fine-Tuned Japanese Riddles.}
\label{tab:japanese_riddles_comparison}
\begin{tabular}{|l|p{6.2cm}|p{6.2cm}|}
\hline
\textbf{Prompting Method} & \textbf{Pretrained Japanese Riddle} & \textbf{Fine-Tuned Japanese Riddle} \\
\hline
Zero-Shot & 頭はあるが泣くことはない床はあるが寝ることはない口はあるが話すことはないそして、変わるが変わらないものなんだ何なのだろう "川" (I have a head, but never weep… A River) & 羽がなくても空を飛び、目がなくても涙を流すものは何？ (“What flies without wings and cries without eyes?”) \\
\hline
Few-Shot & 鍵があるけど、鍵を開けられないものは何？ (What has keys but can't open locks?) & 落とすと割れますが、微笑むと微笑み返します。私は何でしょう？ (“If you drop me, I'm sure to crack; but smile at me, and I'll smile back.”) \\
\hline
Chain-of-Thought & 羽のように軽いのに、最強の男でも一瞬以上は持ちこたえられないものは何でしょう？ (Light as a feather...) & 1分に1度、瞬間に2度、千年に一度も訪れないものは何ですか？ → Mの文字 (“What comes once in a minute, twice in a moment, but never in a thousand years?” → “Letter M”) \\
\hline
Adversarial & 口がないのに話し、耳がないのに聞く。体がないのに風と共に生きる。私は何？ (I speak without a mouth...) & 触れずに壊せるものは何？ (“What can you break without touching it?”) \\
\hline
AOF (Fine-Tuned) & 目には見えず、耳には聞こえず、口には感じないものは何？ (“What can’t be seen, heard, or tasted?”) & 私は音を持たず、光もない。それでも、全てを照らすことができる。 (“I have no sound or light, yet I can illuminate everything.”) \\
\hline
\end{tabular}
\end{table}
\FloatBarrier

\subsection{Japanese Fine-Tuned vs Real-World Riddles}
\begin{table}[H]
\centering
\scriptsize
\caption{Comparison of Real-World vs Fine-Tuned Japanese Riddles.}
\begin{tabular}{|p{6.2cm}|p{6.2cm}|}
\hline
\textbf{Real-World-Style Riddle (EN/JP)} & \textbf{Fine-Tuned-Style Riddle (EN/JP)} \\
\hline
(\textbf{crestecusa.com}) What’s the similarity between the morning newspaper (chōkan: 朝刊) and a Buddhist monk (bōsan: 坊さん)? けさきてきょうよむ(kesa kite kyo yomu) & つかむけど、抱きしめられない。夜にしかできないことは何？夢 (“What can you catch but never hold tight, only in the night? A dream”)  \\ 
\hline
What is the box you can't close once it's opened? (一度開けたらもう戻せない箱は何でしょう？ 記憶 Memory) & たくさん詰まっているけど、何も入れられない袋は何でしょう？ 蜂の巣 (“What is the bag that's full but you can't put anything in it? A beehive”) \\
\hline
\end{tabular}
\label{tab:japanese_riddles_comparison_real_fine_tune}
\end{table}
\FloatBarrier

\end{CJK}
\clearpage
\section{Appendix: Arabic Riddle Examples}
\label{sec:arabic_riddle_examples}

\subsection{Arabic Pretrained Bilingual Riddle Examples}

\clearpage

\subsection{Arabic Pretrained vs Fine-Tuned Bilingual Riddle Examples}

\clearpage

\subsection{Real-World Riddles vs. Fine-Tuned Arabic Riddles}

\clearpage
\section{Appendix: Chinese Riddle Examples}
\label{sec:chinese_riddle_examples}

\begin{CJK}{UTF8}{gbsn}

\subsection{Chinese Pretrained Riddle Examples}
\begin{table}[ht]
\centering
\scriptsize
\caption{Representative Chinese riddles generated under pretrained settings across three models. Each row presents the original riddle in Chinese and English, along with its answer.}
\label{tab:pretrained_riddles_examples_zh}
\begin{tabular}{|c|p{7cm}|p{3cm}|}
\hline
\textbf{Model} & \textbf{Riddle (ZH / EN)} & \textbf{Answer (ZH / EN)} \\
\hline
GPT-4o &
\textbf{ZH:} 口袋里有个圆，白天不见晚上现。 \newline
\textbf{EN:} \textit{There’s a circle in my pocket, unseen by day, revealed at night.} &
\textbf{ZH:} 月亮 \newline
\textbf{EN:} the moon \\
\hline
DeepSeek R1 &
\textbf{ZH:} 身穿白衣不沾尘，举头低垂泪两行。 \newline
\textbf{EN:} \textit{Dressed in white yet never stained, head bowed, two lines of tears descend.} &
\textbf{ZH:} 芦苇 \newline
\textbf{EN:} reed \\
\hline
DeepSeek R1 (alt) &
\textbf{ZH:} 上下两半黄一体，秋风过处伴人归。 \newline
\textbf{EN:} \textit{Two yellow halves joined as one, the autumn breeze leads travelers home.} &
\textbf{ZH:} 稻穗 \newline
\textbf{EN:} rice ear \\
\hline
LLaMA 3.1 &
\textbf{ZH:} 海底无声森林现，触之无枝叶。 \newline
\textbf{EN:} \textit{A silent forest appears beneath the sea; touch it—no branches to see.} &
\textbf{ZH:} 珊瑚 \newline
\textbf{EN:} coral \\
\hline
LLaMA 3.1 (radical) &
\textbf{ZH:} 双人旁上加山石，里边藏着秋波深。 \newline
\textbf{EN:} \textit{With “person” and “mountain rock” radicals, inside lies autumn’s deep ripples.} &
\textbf{ZH:} 留 \newline
\textbf{EN:} the character \textit{liú} \\
\hline
\end{tabular}
\end{table}
\FloatBarrier

\subsection{Chinese Fine-Tuned vs Pretrained Riddle Examples}
\begin{table}[ht]
\centering
\scriptsize
\caption{Chinese fine-tuned GPT-4o riddles compared to pretrained prompts across different methods.}
\label{tab:chinese_finetuned_examples}
\begin{tabular}{|l|p{10cm}|}
\hline
\textbf{Prompting Method} & \textbf{Fine-Tuned GPT-4o Riddle (EN / ZH)} \\
\hline
\textbf{Zero-Shot} & EN: What hides in your pocket by day, yet hangs in the sky by night? \newline ZH: 什么东西，白天躲在口袋里，晚上挂在天上？ \newline \textit{Answer: The moon / 月亮} \\
\hline
\textbf{Few-Shot} & EN: I’m green on the outside, red within, juicy and sweet, a summer win. What am I? \newline ZH: 身穿绿袍，头顶红帽，剥去衣裳，味道真好。 \newline \textit{Answer: Watermelon / 西瓜} \\
\hline
\textbf{Chain-of-Thought} & EN: I can be cracked, made, told, and played. What am I? \newline ZH: 我可以被破解、制造、讲述和玩耍。我是什么？ \newline \textit{Answer: A joke / 笑话} \\
\hline
\textbf{Adversarial} & EN: What goes up but never comes down? \newline ZH: 什么东西只增不减？ \newline \textit{Answer: Age / 年龄} \\
\hline
\textbf{AOF (Ours)} & EN: I run without legs, whisper without a mouth. What am I? \newline ZH: 我无腿而跑，没有嘴却能低语。我是什么？ \newline \textit{Answer: The wind / 风} \\
\hline
\end{tabular}
\end{table}
\FloatBarrier

\subsection{Chinese Fine-Tuned vs Real-World Riddles}
\begin{table}[H]
\centering
\scriptsize
\caption{Chinese riddle comparison: fine-tuned AOF riddles vs real-world 灯谜.}
\label{tab:fine_tuned_realworld_riddles_zh}
\begin{tabular}{|c|p{6.5cm}|p{6.5cm}|}
\hline
\textbf{Row} & \textbf{Real-World 灯谜 (ZH / EN)} & \textbf{AOF Fine-Tuned Riddle (ZH / EN)} \\
\hline
1 & ZH: 口袋里有个圆，白天不见晚上现。 \newline EN: There’s a circle in my pocket, unseen by day, revealed at night. & ZH: 海底藏森林，触之无枝叶，红颜共浪舞，千年不知悔。 \newline EN: A forest hides beneath the sea; touch it—no branch or leaf. Its crimson dances with the waves, unchanged for a thousand years. \\
\hline
\end{tabular}
\end{table}
\FloatBarrier

\subsection{Chinese Fine-Tuned AOF Examples}
\begin{table}[H]
\centering
\scriptsize
\caption{Fine-tuned Chinese riddle examples using AOF prompting.}
\label{tab:chinese_finetuned_riddle_examples}
\begin{tabular}{|c|p{6cm}|p{6cm}|p{2cm}|}
\hline
\textbf{Row} & \textbf{Chinese Riddle} & \textbf{English Translation} & \textbf{Answer} \\
\hline
1 & 口袋里有个圆，白天不见晚上现。 & There’s a circle in my pocket, unseen by day, revealed at night. & 月亮 (Moon) \\
2 & 无声无息钻进来，千言万语藏心怀。 & Silently it slips inside, a thousand words it holds inside. & 信 (Letter) \\
3 & 身穿彩衣，飞舞花丛，白天聚会，晚上无踪… & Dressed in rainbow robes, it dances through the blooms by day…then vanishes by night. & 蝴蝶 (Butterfly) \\
\hline
\end{tabular}
\end{table}
\FloatBarrier

\end{CJK}

\clearpage
\section{Appendix: French Riddle Examples}
\label{sec:french_riddle_examples} 

\subsection{French Pretrained Riddle Examples}
\begin{table}[htbp]
\centering
\scriptsize
\caption{Representative French riddles generated under pretrained settings across three models.}
\label{tab:pretrained_riddles_examples_fr}
\begin{tabular}{|c|p{7cm}|p{3cm}|}
\hline
\textbf{Model} & \textbf{Riddle (FR / EN)} & \textbf{Answer (FR / EN)} \\
\hline
GPT-4o & FR: Je vole sans ailes, je pleure sans yeux… \newline EN: I fly without wings, I cry without eyes… & FR: un nuage \newline EN: a cloud \\
\hline
DeepSeek R1 & FR: J’ai une tête mais je ne pleure jamais… \newline EN: I have a head but never cry… & FR: une rivière \newline EN: a river \\
\hline
LLaMA 3.1 (a) & FR: Je danse sans musique, je ris sans bouche… \newline EN: I dance without music, I laugh without a mouth… & FR: le vent \newline EN: the wind \\
\hline
LLaMA 3.1 (b) & FR: Invisible sur l’écran, je révèle toute l’histoire… \newline EN: Invisible on the screen, I reveal the whole story… & FR: un curseur \newline EN: a cursor \\
\hline
\end{tabular}
\end{table}
\FloatBarrier

\subsection{French Pretrained vs Fine-Tuned}
\begin{table}[htbp]
\centering
\scriptsize
\caption{Comparison of pretrained vs. fine-tuned GPT-4o French riddles across prompting methods.}
\label{tab:French_Riddle_Examples_for_Pretrained_vs_Fine-Tuned}
\begin{tabular}{|l|p{6.5cm}|p{6.5cm}|}
\hline
\textbf{Prompting Method} & \textbf{Pretrained Riddle (EN / FR)} & \textbf{Fine-Tuned Riddle (EN / FR)} \\
\hline
Zero-Shot & EN: I have keys but open no locks… \newline FR: J’ai des clés mais n’ouvre aucun verrou… & EN: What has keys but can’t open a door… \newline FR: Quel est l’objet avec des touches… \\
\hline
Few-Shot & EN: I speak without a mouth and hear without ears… \newline FR: Je parle sans bouche… & EN: I have a neck but no head… \newline FR: J'ai un cou mais pas de tête… \\
\hline
Chain-of-Thought & EN: I can be broken without a sound… \newline FR: Je peux être brisé sans un bruit… & EN: What has keys but can’t open locks… \newline FR: Qu’est-ce qui a des touches mais… \\
\hline
Adversarial & EN: What has keys but can't open locks… \newline FR: Qu'est-ce qui a des clés… & EN: What has keys but can't open locks? \newline FR: Qu'est-ce qui a des clés… \\
\hline
AOF & EN: In the garden of words, I am a bee… \newline FR: Dans le jardin des mots, je suis une abeille… & EN: I slip through fingers like silver and gold… \newline FR: Je glisse entre les doigts… \\
\hline
\end{tabular}
\end{table}
\FloatBarrier

\subsection{French Fine-Tuned Riddles and Their Real-World Counterparts}
\begin{table}[htbp]
\centering
\scriptsize
\caption{French riddle comparison: fine-tuned GPT-4o AOF riddles vs. real-world examples.}
\label{tab:fine_tuned_realworld_riddles_fr}
\begin{tabular}{|c|p{6.5cm}|p{6.5cm}|}
\hline
\textbf{Row} & \textbf{Real-World Riddle} & \textbf{AOF Fine-Tuned Riddle} \\
\hline
1 & FR: Je vole sans ailes, je pleure sans yeux… \newline EN: I fly without wings, I cry without eyes… & FR: Dans le jardin des mots, je suis une abeille… \newline EN: In the garden of words, I am a bee… \\
\hline
\end{tabular}
\end{table}
\FloatBarrier

\subsection{French Fine-Tuned AOF Examples}
\begin{table}[H]
\centering
\scriptsize
\caption{Representative French riddles from the fine-tuned GPT-4o model using AOF.}
\label{tab:french_finetuned_riddle_examples}
\begin{tabular}{|c|p{6.5cm}|p{6.5cm}|}
\hline
\textbf{Row} & \textbf{French Riddle (FR)} & \textbf{English Translation (EN)} \\
\hline
1 & FR: Je disparais au crépuscule, mais je reviens à l'aube. & EN: I disappear at dusk, but return at dawn. \\
\hline
2 & FR: Sur les sols je glisse, ma mission est de nettoyer… & EN: On floors I glide, my mission is to clean… \\
\hline
3 & FR: Je glisse entre les doigts comme l'argent et l'or… & EN: I slip through fingers like silver and gold… \\
\hline
\end{tabular}
\end{table}
\FloatBarrier

\clearpage
\section{Appendix:Prompting Methods}

\label{app:Prompting}

\subsection{Chinese prompts}
\begin{table}[H]
    \centering
    \footnotesize
    \caption{Prompting Methods for English Chinese}
    \begin{tabular}{|p{15cm}|}
        \hline
        \textbf{Zero-Shot Prompting} \\ 
        \texttt{Create 10 bilingual riddle in both Chinese and English. The riddle should be novel, unqiue, clever, engaging, and suitable for all ages.
It should rhyme in English and maintain a poetic or rhythmic flow in Chinese.
The answer should be the same in both languages..} \\ \hline
        
        \textbf{Few-Shot Prompting Example} \\ 
        \texttt{Here are some example riddles:} \\
        \texttt{Riddle: What has keys but can't open locks?} \\
        \texttt{Answer: A piano} \\
        \texttt{Riddle: What has hands but can’t clap?} \\
        \texttt{Answer: A clock} \\
        \texttt{[Riddle Generation Continues...]} \\
        \texttt{Now, generate 10 brand new **bilingual** riddles in **English and Chinese** with **logical wordplay and ambiguity**.} \\ \hline
        
        \textbf{Chain-of-Thought (CoT) Prompting Example} \\ 
        \texttt{Craft 10 clever riddles by reasoning through the following steps:} \\
        \texttt{1. Identify the deeper or metaphorical meanings of the word.} \\
        \texttt{2. Introduce wordplay or ambiguity to mislead or confuse the solver.} \\
        \texttt{3. Add misdirection to guide the reader toward the wrong conclusion.} \\
        \texttt{4. Ensure the riddle remains engaging, poetic, and fun to solve.} \\
        \texttt{5. After the riddle, provide the answer in both English and Chinese, revealing the true meaning.} \\ \hline
        
        \textbf{Adversarial Prompting Example} \\ 
        \texttt{Create 10 tricky creative bilingual riddle in both English and Chinese.
The riddle should intentionally mislead the reader into thinking of one answer
while the correct answer is something unexpected but still logical. Use wordplay,
ambiguity, and misdirection to make the riddle difficult to solve.
The answer must be the same in both languages.} \\ \hline
        
        \textbf{Adaptive Originality Filtering (AOF, Ours) Example } \\ 
        \texttt{Generate 10 completely new bilingual riddles in English and Chinese. Use diverse grammar: poetic, declarative, metaphorical. Avoid repeating openers like `I have'' or I am''. Only 2–3 riddles may end with What am I?''. Others should use endings like ...yet no one remembers me.'' or Still, I linger in the air.'' Avoid common answers such as \{"shadow", "time", "echo", "fire", "breath", "wind", "silence"\}. Chinese versions must match the tone and trickery.} \\ \hline
    \end{tabular}
    \label{tab:prompting_methods}
\end{table}

\clearpage

\subsection{Japanese prompts}
\begin{table}[H]
    \centering
    \footnotesize
    \caption{Prompting Methods for English Japanese}
    \begin{tabular}{|p{15cm}|}
        \hline
        \textbf{Zero-Shot Prompting} \\ 
        \texttt{Create 10 bilingual riddle in both Chinese and English. The riddle should be novel, unqiue, clever, engaging, and suitable for all ages.
It should rhyme in English and maintain a poetic or rhythmic flow in Japanese.
The answer should be the same in both languages..} \\ \hline
        
        \textbf{Few-Shot Prompting Example} \\ 
        \texttt{Here are some example riddles:} \\
        \texttt{Riddle: What has keys but can't open locks?} \\
        \texttt{Answer: A piano} \\
        \texttt{Riddle: What has hands but can’t clap?} \\
        \texttt{Answer: A clock} \\
        \texttt{[Riddle Generation Continues...]} \\
        \texttt{Now, generate 10 brand new **bilingual** riddles in **English and Japanese** with **logical wordplay and ambiguity**.} \\ \hline
        
        \textbf{Chain-of-Thought (CoT) Prompting Example} \\ 
        \texttt{Craft 10 clever riddles by reasoning through the following steps:} \\
        \texttt{1. Identify the deeper or metaphorical meanings of the word.} \\
        \texttt{2. Introduce wordplay or ambiguity to mislead or confuse the solver.} \\
        \texttt{3. Add misdirection to guide the reader toward the wrong conclusion.} \\
        \texttt{4. Ensure the riddle remains engaging, poetic, and fun to solve.} \\
        \texttt{5. After the riddle, provide the answer in both English and Japanese, revealing the true meaning.} \\ \hline
        
        \textbf{Adversarial Prompting Example} \\ 
        \texttt{Create 10 tricky creative bilingual riddle in both English and Japanese.
The riddle should intentionally mislead the reader into thinking of one answer
while the correct answer is something unexpected but still logical. Use wordplay,
ambiguity, and misdirection to make the riddle difficult to solve.
The answer must be the same in both languages.} \\ \hline
        
        \textbf{Adaptive Originality Filtering (AOF, Ours) Example } \\ 
        \texttt{Generate 10 completely new bilingual riddles in English and Japanese. The riddle **must not** be a reworded version of existing riddles. Only 2–3 riddles may end with ``What am I?''. Others should use endings like ``...yet no one remembers me.'' or ``Still, I linger in the air.'' Avoid common answers such as \{"shadow", "time", "echo", "fire", "breath", "wind", "silence"\}. The riddle should be creative, original, and use **unusual objects** or **abstract concept. The riddle **should not** be translated into Japanese from English or change some words} \\ \hline
    \end{tabular}
    \label{tab:prompting_methods_Japanese}
\end{table}

\clearpage
\subsection{Arabic prompts}
\begin{table}[H]
    \centering
    \footnotesize
    \caption{Prompting Methods for English Arabic}
    \begin{tabular}{|p{15cm}|}
        \hline
        \textbf{Zero-Shot Prompting} \\ 
        \texttt{Create 10 bilingual riddle in both Arabic and English. The riddle should be novel, unqiue, clever, engaging, and suitable for all ages.
It should rhyme in English and maintain a poetic or rhythmic flow in Arabic.
The answer should be the same in both languages..} \\ \hline
        
        \textbf{Few-Shot Prompting Example} \\ 
        \texttt{Here are some example riddles:} \\
        \texttt{Riddle: What has keys but can't open locks?} \\
        \texttt{Answer: A piano} \\
        \texttt{Riddle: What has hands but can’t clap?} \\
        \texttt{Answer: A clock} \\
        \texttt{[Riddle Generation Continues...]} \\
        \texttt{Now, generate 10 brand new **bilingual** riddles in **English and Arabic** with **logical wordplay and ambiguity**.} \\ \hline
        
        \textbf{Chain-of-Thought (CoT) Prompting Example} \\ 
        \texttt{Craft 10 clever riddles by reasoning through the following steps:} \\
        \texttt{1. Identify the deeper or metaphorical meanings of the word.} \\
        \texttt{2. Introduce wordplay or ambiguity to mislead or confuse the solver.} \\
        \texttt{3. Add misdirection to guide the reader toward the wrong conclusion.} \\
        \texttt{4. Ensure the riddle remains engaging, poetic, and fun to solve.} \\
        \texttt{5. After the riddle, provide the answer in both English and Arabic, revealing the true meaning.} \\ \hline
        
        \textbf{Adversarial Prompting Example} \\ 
        \texttt{Create 10 tricky creative bilingual riddle in both English and Arabic.
The riddle should intentionally mislead the reader into thinking of one answer
while the correct answer is something unexpected but still logical. Use wordplay,
ambiguity, and misdirection to make the riddle difficult to solve.
The answer must be the same in both languages.} \\ \hline
        
        \textbf{Adaptive Originality Filtering (AOF, Ours) Example } \\ 
        \texttt{Generate 10 completely new bilingual riddles in English and Arabic. Use diverse grammar: poetic, declarative, metaphorical. Avoid repeating openers like ``I have'' or ``I am''. Only 2–3 riddles may end with ``What am I?''. Others should use endings like ``...yet no one remembers me.'' or ``Still, I linger in the air.'' Avoid common answers such as \{"shadow", "time", "echo", "fire", "breath", "wind", "silence"\}. Arabic versions must match the tone and trickery.} \\ \hline
    \end{tabular}
    \label{tab:prompting_methods_Arabic}
\end{table}

\clearpage
\subsection{French prompts}

\begin{table}[H]
    \centering
    \footnotesize
    \caption{Prompting Methods for English French}
    \begin{tabular}{|p{15cm}|}
        \hline
        \textbf{Zero-Shot Prompting} \\ 
        \texttt{Create 10 bilingual riddle in both French and English. The riddle should be novel, unqiue, clever, engaging, and suitable for all ages.
It should rhyme in English and maintain a poetic or rhythmic flow in French.
The answer should be the same in both languages..} \\ \hline
        
        \textbf{Few-Shot Prompting Example} \\ 
        \texttt{Here are some example riddles:} \\
        \texttt{Riddle: What has keys but can't open locks?} \\
        \texttt{Answer: A piano} \\
        \texttt{Riddle: What has hands but can’t clap?} \\
        \texttt{Answer: A clock} \\
        \texttt{[Riddle Generation Continues...]} \\
        \texttt{Now, generate 10 brand new **bilingual** riddles in **English and French** with **logical wordplay and ambiguity**.} \\ \hline
        
        \textbf{Chain-of-Thought (CoT) Prompting Example} \\ 
        \texttt{Craft 10 clever riddles by reasoning through the following steps:} \\
        \texttt{1. Identify the deeper or metaphorical meanings of the word.} \\
        \texttt{2. Introduce wordplay or ambiguity to mislead or confuse the solver.} \\
        \texttt{3. Add misdirection to guide the reader toward the wrong conclusion.} \\
        \texttt{4. Ensure the riddle remains engaging, poetic, and fun to solve.} \\
        \texttt{5. After the riddle, provide the answer in both English and French, revealing the true meaning.} \\ \hline
        
        \textbf{Adversarial Prompting Example} \\ 
        \texttt{Create 10 tricky creative bilingual riddle in both English and French.
The riddle should intentionally mislead the reader into thinking of one answer
while the correct answer is something unexpected but still logical. Use wordplay,
ambiguity, and misdirection to make the riddle difficult to solve.
The answer must be the same in both languages.} \\ \hline
        
        \textbf{Adaptive Originality Filtering (AOF, Ours) Example } \\ 
        \texttt{Generate 10 completely new bilingual riddles in English and French. Use diverse grammar: poetic, declarative, metaphorical. Avoid repeating openers like ``I have'' or ``I am''. Only 2–3 riddles may end with ``What am I?''. Others should use endings like ``...yet no one remembers me.'' or ``Still, I linger in the air.'' Avoid common answers such as \{"shadow", "time", "echo", "fire", "breath", "wind", "silence"\}. French versions must match the tone and trickery.} \\ \hline
    \end{tabular}
    \label{tab:prompting_methods_French}
\end{table}

\clearpage
\section{Appendix: Fined-tuned Training and Evaluation Details}
\label{app:training_details}

\subsection{Dataset Selection and Preparation}
We used the BiRdQA dataset~\citep{zhang2022birdqa}, a multilingual benchmark designed to test figurative language understanding and commonsense inference. It includes 6,614 English riddles and 8,751 Chinese riddles, each paired with four answer options. Riddles were shuffled at each epoch to prevent memorization, and no synthetic augmentation was applied.

Its linguistic diversity—spanning syntactic constructions, cultural idioms, and metaphorical phrasing—made BiRdQA suitable for riddle-based fine-tuning. All data were Unicode-normalized and deduplicated, and stratified sampling ensured balanced language representation.

\subsection{Training Strategy}
Fine-tuning was framed as a supervised multi-class classification problem. The model selected one correct answer out of four using cross-entropy loss. The following hyperparameters were used:

\begin{itemize}
  \item \textbf{Temperature:} 0.7
  \item \textbf{Token Limit:} 3000
  \item \textbf{Initial Accuracy:} 37–59\% on development set
\end{itemize}

Training followed a three-stage pipeline: base fine-tuning, early stopping on dev performance, and multilingual test evaluation to check generalization.

\subsection{Appendix:Training Set Expansion}
To improve abstraction and metaphor handling, the English and Chinese development sets were merged into the training pool. This added examples with closely related distractors and borderline ambiguity. After retraining, test accuracy rose to 97\%.

These improvements suggest the model internalized deep riddle logic, moving beyond surface pattern recognition and toward more sophisticated reasoning involving contradiction and misdirection.

\subsection{Model Comparison Methodology}
\label{app:model_comparison_details}

\subsubsection{Baseline Models}
We benchmarked the fine-tuned GPT-4o against three models:

\begin{itemize}
    \item \textbf{Pretrained GPT-4o (2024-08-06):} Unadapted baseline.
    \item \textbf{LLaMA 3.1:} An open-weight multilingual model with strong reasoning ability.
    \item \textbf{DeepSeek R1:} A reasoning-optimized model focusing on step-wise logical alignment.
\end{itemize}

Each model received the same riddles under consistent prompting strategies to ensure fair comparison.

\subsubsection{Evaluation Procedure}
All models were tested under five prompting strategies (Zero-Shot, Few-Shot, Chain-of-Thought, Adversarial, AOF) with identical templates (Table~\ref{tab:prompting_methods}). Metrics included:

\begin{itemize}
    \item \textbf{Accuracy} (multiple choice prediction)
    \item \textbf{Token Length} (verbosity)
    \item \textbf{Self-BLEU} (semantic diversity)
    \item \textbf{Distinct-2} (lexical uniqueness)
\end{itemize}

Qualitative evaluations by human reviewers assessed metaphor handling, distractor discrimination, and cultural idiomatic fluency.

\subsubsection{Summary of Findings}
Fine-tuned GPT-4o consistently outperformed all baselines across metrics. Key observations:

\begin{itemize}
    \item \textbf{Accuracy:} Rose from 59\% (pretrained) to 97\% (fine-tuned).
    \item \textbf{Reasoning:} Demonstrated superior metaphor resolution and logical contradiction handling.
    \item \textbf{Naturalness:} Generated riddles more closely matched idiomatic structures in both English and Chinese.
\end{itemize}

\subsection{Impact of Multiple-Choice Framing}
\label{app:framing_effects}

Retaining a multiple-choice structure during fine-tuning had a pronounced effect on the model’s ability to reason through ambiguity. Unlike generative formats where any output is valid if semantically relevant, the multiple-choice setup forced the model to:

\begin{itemize}
    \item Distinguish between semantically similar options
    \item Engage in elimination-style reasoning
    \item Learn disambiguation strategies aligned with riddle logic
\end{itemize}

This setup simulated test-like conditions where distractors were deliberately constructed to reflect surface-level similarity (e.g., phonetic overlaps, shared imagery, or logical decoys). The model improved not only in accuracy but in inferential depth.

Moreover, this format likely enhanced the model’s sensitivity to misdirection—a core feature of riddles—by requiring it to reject reasonable but incorrect answers. We observed that this effect carried over to open-ended generation: the model became more likely to embed internal contradiction or layered metaphor, hallmarks of real-world riddles.

In sum, multiple-choice framing served both as a task constraint and as a pedagogical scaffold, encouraging the model to develop strategies beyond rote keyword matching.

\section{Appendix: AOF Prompt Template and Constraints}
\label{app:prompt_structure}

The Adaptive Originality Filtering (AOF) prompt enforces explicit structural rules to maximize diversity, creativity, and cultural fit. Specifically:
\begin{itemize}
  \item \textbf{Syntactic Variety:} At least half of the riddles must use poetic, declarative, or metaphorical forms. Fewer than 3 per batch may end in “What am I?”
  \item \textbf{Answer Filtering:} Outputs with generic answers (e.g., shadow, time, echo, fire, breath) are discarded.
  \item \textbf{Cross-Lingual Parity:} Translations must preserve ambiguity or metaphor across both languages.
  \item \textbf{Novelty Filter:} Semantic similarity to known riddles must fall below a threshold ($\theta = 0.75$), as measured against BiRdQA~\citep{zhang2022birdqa}.
\end{itemize}

\vspace{-0.5em}
\subsection{Semantic Similarity Filtering Equation}
\label{app:semantic_filtering_eq}

A candidate riddle $r_{\text{gen}}$ is compared to a reference dataset $\mathcal{D} = \{r_i\}_{i=1}^N$ via:

\begin{equation}
\mathcal{S}(r_\text{gen}, \mathcal{D}) = \max_{r_i \in \mathcal{D}} \cos(\phi(r_\text{gen}), \phi(r_i))
\end{equation}

\noindent where $\phi(\cdot)$ is an embedding function (e.g., \texttt{all-MiniLM-L6-v2}). A candidate passes if $\mathcal{S} < \theta = 0.75$.

\vspace{-0.5em}
\subsection{Rejection Sampling Algorithm}
\label{app:rejection_algo}

\begin{algorithm}[H]
\caption{AOF Rejection Sampling}
\begin{algorithmic}[1]
\State \textbf{Input:} Prompt $P$, Model $M$, Reference Set $\mathcal{D}$, Threshold $\theta$, MaxAttempts $k$
\For{$j = 1$ to $k$}
    \State $r_\text{gen} \leftarrow M(P)$
    \State $\mathcal{S} \leftarrow \max_{r_i \in \mathcal{D}} \cos(\phi(r_\text{gen}), \phi(r_i))$
    \If{$\mathcal{S} < \theta$}
        \State \textbf{return} $r_\text{gen}$
    \EndIf
\EndFor
\State \textbf{return} None
\end{algorithmic}
\end{algorithm}

\vspace{-0.5em}
\subsection{Threshold Sensitivity: Self-BLEU and Distinct-2}
\label{app:threshold_sensitivity}

Table~\ref{tab:threshold_sensitivity_bleu_distinct} shows how Self-BLEU and Distinct-2 vary under different novelty thresholds ($\theta$) for three models. The optimal balance of diversity and non-redundancy appears at $\theta=0.75$ for all models.

\vspace{0.5em}
\begin{table*}[ht]
\centering
\small
\caption{
\textbf{Self-BLEU and Distinct-2 at different novelty thresholds $\theta$ across models on English–Chinese.} 
Lower Self-BLEU and higher Distinct-2 reflect better originality and lexical diversity.
}
\label{tab:threshold_sensitivity_bleu_distinct}
\begin{tabular}{|l|c|c|c|c|}
\hline
\textbf{Language} & \textbf{Model} & \textbf{Threshold $\theta$} & \textbf{Self-BLEU} & \textbf{Distinct-2} \\
\hline
\multirow{4}{*}{English–Chinese} 
& GPT-4o & 0.65 & 0.231 & 0.649 \\
&        & 0.70 & 0.311 & 0.846 \\
&        & \textbf{0.75} & \textbf{0.280} & \textbf{0.869} \\
&        & 0.80 & 0.434 & 0.824 \\
\hline
\multirow{4}{*}{English–Chinese} 
& LLaMA 3.1 & 0.65 & 0.577 & 0.621 \\
&           & 0.70 & 0.573 & 0.826 \\
&           & \textbf{0.75} & \textbf{0.428} & \textbf{0.776} \\
&           & 0.80 & 0.655 & 0.634 \\
\hline
\multirow{4}{*}{English–Chinese} 
& DeepSeek R1 & 0.65 & 0.610 & 0.600 \\
&             & 0.70 & 0.482 & 0.793 \\
&             & \textbf{0.75} & \textbf{0.433} & \textbf{0.674} \\
&             & 0.80 & 0.523 & 0.628 \\
\hline
\end{tabular}
\end{table*}

\section{Appendix: Experimental Configuration Details}
\label{app:appendix_experiments}

\paragraph{Models We evaluated:}
\begin{itemize}
  \item \textbf{GPT-4o (OpenAI)}: Proprietary multilingual model optimized for reasoning and conversational tasks.
  \item \textbf{LLaMA 3.1 (Meta)}: Open-weight transformer trained on internet-scale corpora.
  \item \textbf{DeepSeek Reasoning (R1)}: Fine-tuned for multilingual logical inference.
\end{itemize}
All models were accessed via API with uniform generation parameters: temperature = 0.7 and max token length = 3000.

\paragraph{Prompting Strategies.} We compared:
\begin{itemize}
  \item \textbf{Zero-Shot}: Instruction-only prompting with no exemplars.
  \item \textbf{Few-Shot}: 3–5 riddle-answer pairs per prompt.
  \item \textbf{Chain-of-Thought (CoT)}: Intermediate reasoning steps added to facilitate abstraction.
  \item \textbf{Adversarial}: Distractor-rich prompts based on known LLM vulnerabilities~\citep{wallace2019universal, ribeiro2018semantically}.
  \item \textbf{Adaptive Originality Filtering (AOF)}: Filtering-based prompting for semantic novelty. See  Appendix~\ref{app:prompt_structure}.
\end{itemize}

Prompt formatting logic appears in Appendix~\ref{app:Prompting}

\paragraph{Dataset.} We used BiRdQA~\citep{zhang2022birdqa}, which contains:
\begin{itemize}
  \item 6,614 riddles in English and 8,751 in Chinese.
  \item Multiple-choice format with 1 correct answer and 4 distractors.
\end{itemize}
Few-shot exemplars and semantic filters were drawn from the training splits.

\paragraph{Evaluation Metrics.} We used:
\begin{itemize}
  \item \textbf{Self-BLEU (n=2)}: Measures inter-riddle redundancy. Lower = better.
  \item \textbf{Distinct-2}: Measures lexical diversity via bigram ratios. Higher = better.
  \item \textbf{Cross-lingual BERTScore}: Captures semantic similarity between translations.
  \item \textbf{Syntactic Validity}: Uses spaCy (English/French) and Stanza (Chinese, Arabic, Japanese) to validate parse trees.
  \item \textbf{RiddleScore}: Our composite metric combining novelty, fluency, and alignment.
\end{itemize}

\section{RiddleScore: Implementation and Weight Ablation}
\label{app:riddlescore}

\subsection{Component Formulations}
\textbf{Novelty} (1--max cosine), \textbf{Diversity} (Distinct-2),
\textbf{Fluency} (1/(1+PPL)), and \textbf{Alignment} (BERTScore) follow the
definitions in the main text. All scores are linearly scaled to $[0,1]$.

\paragraph{Why these back-end models?}
We adopt lightweight yet well-validated checkpoints for each sub-metric:

\begin{itemize}
    \item \textbf{MiniLM (all-MiniLM-L6-v2) for Novelty.}  
    MiniLM approaches BERT’s semantic accuracy while running $\sim\!6\times$ faster and using under half the parameters, an ideal trade-off for large-scale cosine filtering~\citep{wang2020minilm}.
    
    \item \textbf{Distinct-2 for Diversity.}  
    This token-level ratio, introduced by \citet{li2016diversity}, remains the de-facto measure of lexical variety and correlates with human “interestingness” ratings in dialogue generation studies.
    
    \item \textbf{GPT-2.5 perplexity for Fluency.}  
    GPT-2.5 PPL shows the strongest alignment with human fluency scores in the HumEval survey of style-transfer metrics~\citep{lai2022human}, and is reference-free and language-agnostic.
    
    \item \textbf{BERTScore for Alignment.}  
    Across 363 MT/captioning systems, BERTScore yields the highest system-level correlation with human adequacy in the ICLR-2020 large-scale evaluation~\citep{zhang2019bertscore}. We employ language-specific checkpoints to avoid cross-lingual degradation noted by later work.
\end{itemize}

Together, these models provide a strong speed--accuracy balance and documented
human-alignment advantages, justifying their use in \textsc{RiddleScore}.

\begin{table}[h]
\centering\small
\begin{tabular}{cccc|c}
\toprule
$\alpha$ & $\beta$ & $\gamma$ & $\delta$ & $\rho$ \\
\midrule
0.25 & 0.25 & 0.25 & 0.25 & 0.71 \\
\textbf{0.30} & \textbf{0.20} & \textbf{0.30} & \textbf{0.20} & \textbf{0.83} \\
0.35 & 0.15 & 0.30 & 0.20 & 0.80 \\
\bottomrule
\end{tabular}
\caption{Spearman correlation with human scores for representative weight
settings (best in bold).}
\label{app:riddlescore_weight_ablation}
\end{table}

\vspace{-2pt}
\noindent
This ablation confirms that slightly heavier emphasis on \textsc{Novelty} and
\textsc{Fluency} best aligns with human judgments of riddle quality.

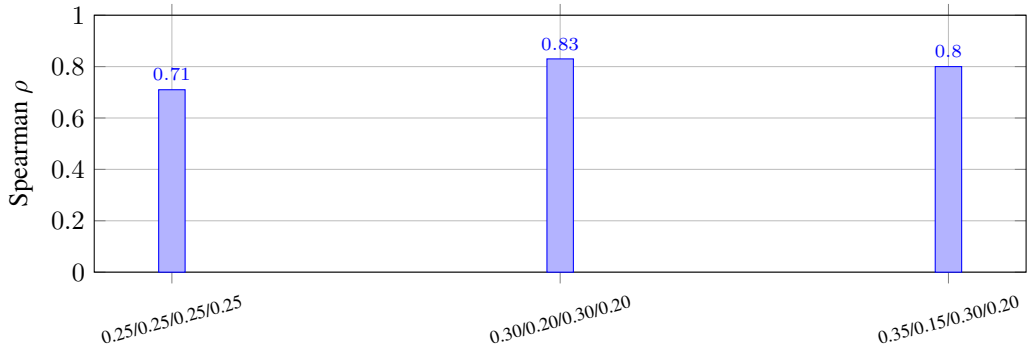
\begin{figure}[t]
\centering
\begin{tikzpicture}
\begin{axis}[
    ybar,
    bar width=10pt,
    width=\columnwidth, 
    height=5cm,
    ymin=0, ymax=1,
    ylabel={Spearman $\rho$},
    symbolic x coords={0.25/0.25/0.25/0.25, 0.30/0.20/0.30/0.20, 0.35/0.15/0.30/0.20},
    xtick=data,
    xticklabel style={font=\scriptsize, align=center, rotate=15},
    nodes near coords,
    nodes near coords align={vertical},
    every node near coord/.append style={font=\scriptsize},
    grid=both
]
\addplot coordinates {
    (0.25/0.25/0.25/0.25,0.71)
    (0.30/0.20/0.30/0.20,0.83)
    (0.35/0.15/0.30/0.20,0.80)
};
\end{axis}
\end{tikzpicture}
\caption{
Spearman correlation between RiddleScore and human ratings under different weight settings. Higher $\rho$ indicates stronger alignment.
}
\label{fig:weight_ablation}
\end{figure}

\section{Appendix: Human Annotation Design and Rationale}
\label{app:human_annotation}

To supplement automatic evaluation, we developed a four-part human annotation rubric, presented in Table~\ref{tab:human_rubric_fine_tuned} and Table~\ref{tab:human_rubric_pretrained}, to assess the quality of model-generated riddles across languages. Below, we outline the rationale and supporting research for each criterion.

\paragraph{Fluency.}
We assess fluency as the degree to which the riddle adheres to the grammar, syntax, and idiomatic expressions of the target language. This follows standard practices in NLG evaluation where fluency serves as a proxy for readability and linguistic naturalness~\citep{cahill2009nlgmetrics,van2018fluency}.

\paragraph{Novelty.}
Novelty is a measure of how creatively the riddle diverges from common or memorized structures. Annotators are instructed to penalize riddles that resemble known examples or rote templates. Prior work on evaluating creativity in language models emphasizes the importance of semantic originality and variation in structure~\citep{dang2022evaluating,van2019assessing}.

\paragraph{Cultural Fit.}
This dimension captures how well a riddle respects linguistic or cultural norms (e.g., appropriate metaphors, poetic forms, or idiomatic references). For multilingual riddle generation, cultural grounding is essential~\citep{ponti2020xcopa,peng2023language}, especially when metaphoric reasoning is tied to local symbolism or oral traditions~\citep{lakoff1980metaphors}.

\paragraph{Answerability.}
Inspired by QA evaluation practices, we define answerability as the logical coherence between the riddle and its answer. This aligns with the criterion of “solvability” often applied in linguistic humor and riddle literature~\citep{koestler1964act,attardo1994linguistic}, ensuring that riddles are not only poetic but cognitively tractable.

\paragraph{Scoring Procedure.}
Each criterion is rated on a 5-point Likert scale. Annotators were trained using a short calibration phase with real-world riddles from the BiRdQA corpus~\citep{zhang2022birdqa}. Disagreements were resolved by averaging multiple ratings per item, following best practices in subjective NLG evaluation~\citep{van2019assessing}.

\clearpage
\subsection{Human Evaluation Rubric for Pretrained Models}
\begin{table}[H]
\centering
\scriptsize
\caption{Human evaluation rubric for assessing cultural and linguistic preservation in \textbf{pretrained models}.}
\label{tab:human_rubric_pretrained}
\begin{tabular}{|l|p{10cm}|}
\hline
\textbf{Dimension} & \textbf{Evaluation Criteria} \\
\hline
Cultural and Linguistic Preservation & Prompting methods evaluated: Zero-Shot, Few-Shot, Chain-of-Thought, Adversarial, Adaptive Originality Filtering (AOF). Question: “How well does each prompting method preserve cultural and linguistic characteristics in its riddles?” Aspects considered: idioms, metaphor styles, poetic forms, humor, puns, cultural references. Rating scale: 1 = Very Poor, 2 = Poor, 3 = Moderate, 4 = Good, 5 = Excellent. \\
\hline
Free-Response Feedback & “Which prompting method produced the least effective riddles? Why?” “Which prompting method produced the most effective riddles? Why?” \\
\hline
\end{tabular}
\end{table}

\subsection{Human Evaluation Rubric for Fine-Tuned Models}
\begin{table}[H]
\centering
\scriptsize
\caption{Human evaluation rubric for assessing cultural and linguistic preservation in \textbf{fine-tuned models}.}
\label{tab:human_rubric_fine_tuned}
\begin{tabular}{|l|p{10cm}|}
\hline
\textbf{Dimension} & \textbf{Evaluation Criteria} \\
\hline
Cultural and Linguistic Preservation & Prompting methods evaluated: Zero-Shot, Few-Shot, Chain-of-Thought, Adversarial, Adaptive Originality Filtering (AOF). Question: “How well does each prompting method preserve cultural and linguistic characteristics in its riddles?” Aspects considered: idioms, metaphor styles, poetic forms, humor, puns, cultural references. Rating scale: 1 = Very Poor, 2 = Poor, 3 = Moderate, 4 = Good, 5 = Excellent. \\
\hline
Free-Response Feedback & “Which prompting method produced the least effective riddles? Why?” “Which prompting method produced the most effective riddles? Why?” \\
\hline
\end{tabular}
\end{table}


\newpage
\section*{NeurIPS Paper Checklist}

The checklist is designed to encourage best practices for responsible machine learning research, addressing issues of reproducibility, transparency, research ethics, and societal impact. Do not remove the checklist: {\bf The papers not including the checklist will be desk rejected.} The checklist should follow the references and follow the (optional) supplemental material.  The checklist does NOT count towards the page
limit. 

Please read the checklist guidelines carefully for information on how to answer these questions. For each question in the checklist:
\begin{itemize}
    \item You should answer \answerYes{}, \answerNo{}, or \answerNA{}.
    \item \answerNA{} means either that the question is Not Applicable for that particular paper or the relevant information is Not Available.
    \item Please provide a short (1–2 sentence) justification right after your answer (even for NA). 
\end{itemize}

{\bf The checklist answers are an integral part of your paper submission.} They are visible to the reviewers, area chairs, senior area chairs, and ethics reviewers. You will be asked to also include it (after eventual revisions) with the final version of your paper, and its final version will be published with the paper.

The reviewers of your paper will be asked to use the checklist as one of the factors in their evaluation. While "\answerYes{}" is generally preferable to "\answerNo{}", it is perfectly acceptable to answer "\answerNo{}" provided a proper justification is given (e.g., "error bars are not reported because it would be too computationally expensive" or "we were unable to find the license for the dataset we used"). In general, answering "\answerNo{}" or "\answerNA{}" is not grounds for rejection. While the questions are phrased in a binary way, we acknowledge that the true answer is often more nuanced, so please just use your best judgment and write a justification to elaborate. All supporting evidence can appear either in the main paper or the supplemental material, provided in appendix. If you answer \answerYes{} to a question, in the justification please point to the section(s) where related material for the question can be found.

IMPORTANT, please:
\begin{itemize}
    \item {\bf Delete this instruction block, but keep the section heading ``NeurIPS Paper Checklist"},
    \item  {\bf Keep the checklist subsection headings, questions/answers and guidelines below.}
    \item {\bf Do not modify the questions and only use the provided macros for your answers}.
\end{itemize}


\begin{enumerate}

\item {\bf Claims}
    \item[] Question: Do the main claims made in the abstract and introduction accurately reflect the paper's contributions and scope?
    \item[] Answer: \answerYes{} 
    \item[] Justification: The abstract and introduction clearly state our contributions—Adaptive Originality Filtering (AOF) and RiddleScore—and these are directly supported by our experimental results and human evaluations. 
    \item[] Guidelines:
    \begin{itemize}
        \item The answer NA means that the abstract and introduction do not include the claims made in the paper.
        \item The abstract and/or introduction should clearly state the claims made, including the contributions made in the paper and important assumptions and limitations. A No or NA answer to this question will not be perceived well by the reviewers. 
        \item The claims made should match theoretical and experimental results, and reflect how much the results can be expected to generalize to other settings. 
        \item It is fine to include aspirational goals as motivation as long as it is clear that these goals are not attained by the paper. 
    \end{itemize}

\item {\bf Limitations}
    \item[] Question: Does the paper discuss the limitations of the work performed by the authors?
    \item[] Answer: \answerYes{} 
    \item[] Justification: We provide a dedicated Limitations section after the Conclusion that discusses dataset scope, potential weaknesses of our filtering method, and the limited scale of human evaluation.
    \item[] Guidelines:
    \begin{itemize}
        \item The answer NA means that the paper has no limitation while the answer No means that the paper has limitations, but those are not discussed in the paper. 
        \item The authors are encouraged to create a separate "Limitations" section in their paper.
        \item The paper should point out any strong assumptions and how robust the results are to violations of these assumptions (e.g., independence assumptions, noiseless settings, model well-specification, asymptotic approximations only holding locally). The authors should reflect on how these assumptions might be violated in practice and what the implications would be.
        \item The authors should reflect on the scope of the claims made, e.g., if the approach was only tested on a few datasets or with a few runs. In general, empirical results often depend on implicit assumptions, which should be articulated.
        \item The authors should reflect on the factors that influence the performance of the approach. For example, a facial recognition algorithm may perform poorly when image resolution is low or images are taken in low lighting. Or a speech-to-text system might not be used reliably to provide closed captions for online lectures because it fails to handle technical jargon.
        \item The authors should discuss the computational efficiency of the proposed algorithms and how they scale with dataset size.
        \item If applicable, the authors should discuss possible limitations of their approach to address problems of privacy and fairness.
        \item While the authors might fear that complete honesty about limitations might be used by reviewers as grounds for rejection, a worse outcome might be that reviewers discover limitations that aren't acknowledged in the paper. The authors should use their best judgment and recognize that individual actions in favor of transparency play an important role in developing norms that preserve the integrity of the community. Reviewers will be specifically instructed to not penalize honesty concerning limitations.
    \end{itemize}

\item {\bf Theory assumptions and proofs}
    \item[] Question: For each theoretical result, does the paper provide the full set of assumptions and a complete (and correct) proof?
    \item[] Answer: \answerNA{} 
    \item[] Justification: Our paper does not include formal theoretical results or proofs, as it is primarily empirical and methodological in nature.
    \item[] Guidelines:
    \begin{itemize}
        \item The answer NA means that the paper does not include theoretical results. 
        \item All the theorems, formulas, and proofs in the paper should be numbered and cross-referenced.
        \item All assumptions should be clearly stated or referenced in the statement of any theorems.
        \item The proofs can either appear in the main paper or the supplemental material, but if they appear in the supplemental material, the authors are encouraged to provide a short proof sketch to provide intuition. 
        \item Inversely, any informal proof provided in the core of the paper should be complemented by formal proofs provided in appendix or supplemental material.
        \item Theorems and Lemmas that the proof relies upon should be properly referenced. 
    \end{itemize}

    \item {\bf Experimental result reproducibility}
    \item[] Question: Does the paper fully disclose all the information needed to reproduce the main experimental results of the paper to the extent that it affects the main claims and/or conclusions of the paper (regardless of whether the code and data are provided or not)?
    \item[] Answer: \answerYes{} 
    \item[] Justification: We include all prompt templates, the full rejection-sampling loop, metric definitions, and references to the BiRdQA dataset, which together make reproduction of our results possible.
    \item[] Guidelines:
    \begin{itemize}
        \item The answer NA means that the paper does not include experiments.
        \item If the paper includes experiments, a No answer to this question will not be perceived well by the reviewers: Making the paper reproducible is important, regardless of whether the code and data are provided or not.
        \item If the contribution is a dataset and/or model, the authors should describe the steps taken to make their results reproducible or verifiable. 
        \item Depending on the contribution, reproducibility can be accomplished in various ways. For example, if the contribution is a novel architecture, describing the architecture fully might suffice, or if the contribution is a specific model and empirical evaluation, it may be necessary to either make it possible for others to replicate the model with the same dataset, or provide access to the model. In general. releasing code and data is often one good way to accomplish this, but reproducibility can also be provided via detailed instructions for how to replicate the results, access to a hosted model (e.g., in the case of a large language model), releasing of a model checkpoint, or other means that are appropriate to the research performed.
        \item While NeurIPS does not require releasing code, the conference does require all submissions to provide some reasonable avenue for reproducibility, which may depend on the nature of the contribution. For example
        \begin{enumerate}
            \item If the contribution is primarily a new algorithm, the paper should make it clear how to reproduce that algorithm.
            \item If the contribution is primarily a new model architecture, the paper should describe the architecture clearly and fully.
            \item If the contribution is a new model (e.g., a large language model), then there should either be a way to access this model for reproducing the results or a way to reproduce the model (e.g., with an open-source dataset or instructions for how to construct the dataset).
            \item We recognize that reproducibility may be tricky in some cases, in which case authors are welcome to describe the particular way they provide for reproducibility. In the case of closed-source models, it may be that access to the model is limited in some way (e.g., to registered users), but it should be possible for other researchers to have some path to reproducing or verifying the results.
        \end{enumerate}
    \end{itemize}

\item {\bf Open access to data and code}
    \item[] Question: Does the paper provide open access to the data and code, with sufficient instructions to faithfully reproduce the main experimental results, as described in supplemental material?
    \item[] Answer: \answerNo{} 
    \item[] Justification: While the BiRdQA dataset we use is publicly available, we do not provide open-source code at submission time due to anonymity requirements, though all reproduction details are included in the paper and appendix.
    \item[] Guidelines:
    \begin{itemize}
        \item The answer NA means that paper does not include experiments requiring code.
        \item Please see the NeurIPS code and data submission guidelines (\url{https://nips.cc/public/guides/CodeSubmissionPolicy}) for more details.
        \item While we encourage the release of code and data, we understand that this might not be possible, so “No” is an acceptable answer. Papers cannot be rejected simply for not including code, unless this is central to the contribution (e.g., for a new open-source benchmark).
        \item The instructions should contain the exact command and environment needed to run to reproduce the results. See the NeurIPS code and data submission guidelines (\url{https://nips.cc/public/guides/CodeSubmissionPolicy}) for more details.
        \item The authors should provide instructions on data access and preparation, including how to access the raw data, preprocessed data, intermediate data, and generated data, etc.
        \item The authors should provide scripts to reproduce all experimental results for the new proposed method and baselines. If only a subset of experiments are reproducible, they should state which ones are omitted from the script and why.
        \item At submission time, to preserve anonymity, the authors should release anonymized versions (if applicable).
        \item Providing as much information as possible in supplemental material (appended to the paper) is recommended, but including URLs to data and code is permitted.
    \end{itemize}

\item {\bf Experimental setting/details}
    \item[] Question: Does the paper specify all the training and test details (e.g., data splits, hyperparameters, how they were chosen, type of optimizer, etc.) necessary to understand the results?
    \item[] Answer: \answerYes{} 
    \item[] Justification: We specify the datasets, models, prompting strategies, hyperparameters, and evaluation metrics in the Experimental Setup section.
    \item[] Guidelines:
    \begin{itemize}
        \item The answer NA means that the paper does not include experiments.
        \item The experimental setting should be presented in the core of the paper to a level of detail that is necessary to appreciate the results and make sense of them.
        \item The full details can be provided either with the code, in appendix, or as supplemental material.
    \end{itemize}

\item {\bf Experiment statistical significance}
    \item[] Question: Does the paper report error bars suitably and correctly defined or other appropriate information about the statistical significance of the experiments?
    \item[] Answer: \answerYes{} 
    \item[] Justification: We report percentage improvements and Spearman correlation with human judgments, as well as a sensitivity analysis of the novelty threshold, to establish statistical reliability.
    \item[] Guidelines:
    \begin{itemize}
        \item The answer NA means that the paper does not include experiments.
        \item The authors should answer "Yes" if the results are accompanied by error bars, confidence intervals, or statistical significance tests, at least for the experiments that support the main claims of the paper.
        \item The factors of variability that the error bars are capturing should be clearly stated (for example, train/test split, initialization, random drawing of some parameter, or overall run with given experimental conditions).
        \item The method for calculating the error bars should be explained (closed form formula, call to a library function, bootstrap, etc.)
        \item The assumptions made should be given (e.g., Normally distributed errors).
        \item It should be clear whether the error bar is the standard deviation or the standard error of the mean.
        \item It is OK to report 1-sigma error bars, but one should state it. The authors should preferably report a 2-sigma error bar than state that they have a 96\% CI, if the hypothesis of Normality of errors is not verified.
        \item For asymmetric distributions, the authors should be careful not to show in tables or figures symmetric error bars that would yield results that are out of range (e.g. negative error rates).
        \item If error bars are reported in tables or plots, The authors should explain in the text how they were calculated and reference the corresponding figures or tables in the text.
    \end{itemize}

\item {\bf Experiments compute resources}
    \item[] Question: For each experiment, does the paper provide sufficient information on the computer resources (type of compute workers, memory, time of execution) needed to reproduce the experiments?
    \item[] Answer: \answerYes{} 
    \item[] Justification: Our experiments were conducted using hosted LLM APIs (GPT-4o, LLaMA 3.1, and DeepSeek) rather than local GPUs or CPUs. Since model training and inference were performed through provider infrastructure, no special compute hardware details are required beyond the description of models and prompting setup already included in the Experimental Setup.
    \item[] Guidelines:
    \begin{itemize}
        \item The answer NA means that the paper does not include experiments.
        \item The paper should indicate the type of compute workers CPU or GPU, internal cluster, or cloud provider, including relevant memory and storage.
        \item The paper should provide the amount of compute required for each of the individual experimental runs as well as estimate the total compute. 
        \item The paper should disclose whether the full research project required more compute than the experiments reported in the paper (e.g., preliminary or failed experiments that didn't make it into the paper). 
    \end{itemize}
    
\item {\bf Code of ethics}
    \item[] Question: Does the research conducted in the paper conform, in every respect, with the NeurIPS Code of Ethics \url{https://neurips.cc/public/EthicsGuidelines}?
    \item[] Answer: \answerYes{} 
    \item[] Justification: Our Ethics Statement discusses language equity, cultural representation, originality risks, and responsible fine-tuning, ensuring compliance with the NeurIPS Code of Ethics.
    \item[] Guidelines:
    \begin{itemize}
        \item The answer NA means that the authors have not reviewed the NeurIPS Code of Ethics.
        \item If the authors answer No, they should explain the special circumstances that require a deviation from the Code of Ethics.
        \item The authors should make sure to preserve anonymity (e.g., if there is a special consideration due to laws or regulations in their jurisdiction).
    \end{itemize}

\item {\bf Broader impacts}
    \item[] Question: Does the paper discuss both potential positive societal impacts and negative societal impacts of the work performed?
    \item[] Answer: \answerYes{} 
    \item[] Justification: We discuss potential positive impacts of culturally grounded generation and note risks such as possible misuse for misinformation in the Ethics Statement.
    \item[] Guidelines:
    \begin{itemize}
        \item The answer NA means that there is no societal impact of the work performed.
        \item If the authors answer NA or No, they should explain why their work has no societal impact or why the paper does not address societal impact.
        \item Examples of negative societal impacts include potential malicious or unintended uses (e.g., disinformation, generating fake profiles, surveillance), fairness considerations (e.g., deployment of technologies that could make decisions that unfairly impact specific groups), privacy considerations, and security considerations.
        \item The conference expects that many papers will be foundational research and not tied to particular applications, let alone deployments. However, if there is a direct path to any negative applications, the authors should point it out. For example, it is legitimate to point out that an improvement in the quality of generative models could be used to generate deepfakes for disinformation. On the other hand, it is not needed to point out that a generic algorithm for optimizing neural networks could enable people to train models that generate Deepfakes faster.
        \item The authors should consider possible harms that could arise when the technology is being used as intended and functioning correctly, harms that could arise when the technology is being used as intended but gives incorrect results, and harms following from (intentional or unintentional) misuse of the technology.
        \item If there are negative societal impacts, the authors could also discuss possible mitigation strategies (e.g., gated release of models, providing defenses in addition to attacks, mechanisms for monitoring misuse, mechanisms to monitor how a system learns from feedback over time, improving the efficiency and accessibility of ML).
    \end{itemize}
    
\item {\bf Safeguards}
    \item[] Question: Does the paper describe safeguards that have been put in place for responsible release of data or models that have a high risk for misuse (e.g., pretrained language models, image generators, or scraped datasets)?
    \item[] Answer: \answerNA{} 
    \item[] Justification: We do not release any new models or datasets with high misuse risk, so safeguards are not applicable.
    \item[] Guidelines:
    \begin{itemize}
        \item The answer NA means that the paper poses no such risks.
        \item Released models that have a high risk for misuse or dual-use should be released with necessary safeguards to allow for controlled use of the model, for example by requiring that users adhere to usage guidelines or restrictions to access the model or implementing safety filters. 
        \item Datasets that have been scraped from the Internet could pose safety risks. The authors should describe how they avoided releasing unsafe images.
        \item We recognize that providing effective safeguards is challenging, and many papers do not require this, but we encourage authors to take this into account and make a best faith effort.
    \end{itemize}

\item {\bf Licenses for existing assets}
    \item[] Question: Are the creators or original owners of assets (e.g., code, data, models), used in the paper, properly credited and are the license and terms of use explicitly mentioned and properly respected?
    \item[] Answer: \answerYes{} 
    \item[] Justification: We properly cite the BiRdQA dataset and other prior works and models, ensuring that all existing assets are credited and used under their terms.
    \item[] Guidelines:
    \begin{itemize}
        \item The answer NA means that the paper does not use existing assets.
        \item The authors should cite the original paper that produced the code package or dataset.
        \item The authors should state which version of the asset is used and, if possible, include a URL.
        \item The name of the license (e.g., CC-BY 4.0) should be included for each asset.
        \item For scraped data from a particular source (e.g., website), the copyright and terms of service of that source should be provided.
        \item If assets are released, the license, copyright information, and terms of use in the package should be provided. For popular datasets, \url{paperswithcode.com/datasets} has curated licenses for some datasets. Their licensing guide can help determine the license of a dataset.
        \item For existing datasets that are re-packaged, both the original license and the license of the derived asset (if it has changed) should be provided.
        \item If this information is not available online, the authors are encouraged to reach out to the asset's creators.
    \end{itemize}

\item {\bf New assets}
    \item[] Question: Are new assets introduced in the paper well documented and is the documentation provided alongside the assets?
    \item[] Answer: \answerNA{} 
    \item[] Justification: We do not introduce new datasets or models in this paper, so this item does not apply.
    \item[] Guidelines:
    \begin{itemize}
        \item The answer NA means that the paper does not release new assets.
        \item Researchers should communicate the details of the dataset/code/model as part of their submissions via structured templates. This includes details about training, license, limitations, etc. 
        \item The paper should discuss whether and how consent was obtained from people whose asset is used.
        \item At submission time, remember to anonymize your assets (if applicable). You can either create an anonymized URL or include an anonymized zip file.
    \end{itemize}

\item {\bf Crowdsourcing and research with human subjects}
    \item[] Question: For crowdsourcing experiments and research with human subjects, does the paper include the full text of instructions given to participants and screenshots, if applicable, as well as details about compensation (if any)? 
    \item[] Answer: \answerYes{} 
    \item[] Justification: Our human evaluation involved native speakers following standardized rubrics, and we describe the evaluation process in Appendix P.
    \item[] Guidelines:
    \begin{itemize}
        \item The answer NA means that the paper does not involve crowdsourcing nor research with human subjects.
        \item Including this information in the supplemental material is fine, but if the main contribution of the paper involves human subjects, then as much detail as possible should be included in the main paper. 
        \item According to the NeurIPS Code of Ethics, workers involved in data collection, curation, or other labor should be paid at least the minimum wage in the country of the data collector. 
    \end{itemize}

\item {\bf Institutional review board (IRB) approvals or equivalent for research with human subjects}
    \item[] Question: Does the paper describe potential risks incurred by study participants, whether such risks were disclosed to the subjects, and whether Institutional Review Board (IRB) approvals (or an equivalent approval/review based on the requirements of your country or institution) were obtained?
    \item[] Answer: \answerNA{} 
    \item[] Justification: Our small-scale human evaluation did not involve sensitive data or risks requiring IRB approval, so this item does not apply.
    \item[] Guidelines:
    \begin{itemize}
        \item The answer NA means that the paper does not involve crowdsourcing nor research with human subjects.
        \item Depending on the country in which research is conducted, IRB approval (or equivalent) may be required for any human subjects research. If you obtained IRB approval, you should clearly state this in the paper. 
        \item We recognize that the procedures for this may vary significantly between institutions and locations, and we expect authors to adhere to the NeurIPS Code of Ethics and the guidelines for their institution. 
        \item For initial submissions, do not include any information that would break anonymity (if applicable), such as the institution conducting the review.
    \end{itemize}

\item {\bf Declaration of LLM usage}
    \item[] Question: Does the paper describe the usage of LLMs if it is an important, original, or non-standard component of the core methods in this research? Note that if the LLM is used only for writing, editing, or formatting purposes and does not impact the core methodology, scientific rigorousness, or originality of the research, declaration is not required.
    \item[] Answer: \answerYes{}{} 
    \item[] Justification: We clearly describe our use of GPT-4o, LLaMA, and DeepSeek as core components for generation and evaluation, including both pretrained and fine-tuned usage.
    \item[] Guidelines:
    \begin{itemize}
        \item The answer NA means that the core method development in this research does not involve LLMs as any important, original, or non-standard components.
        \item Please refer to our LLM policy (\url{https://neurips.cc/Conferences/2025/LLM}) for what should or should not be described.
    \end{itemize}

\end{enumerate}

\end{document}